\documentclass[sigconf, screen]{acmart}
\usepackage{amsmath,amsfonts,bm}

\def\eqref#1{equation~\ref{#1}}

\def\1{\bm{1}}

\DeclareMathAlphabet{\mathsfit}{\encodingdefault}{\sfdefault}{m}{sl}
\SetMathAlphabet{\mathsfit}{bold}{\encodingdefault}{\sfdefault}{bx}{n}

\usepackage{hyperref}
\usepackage{url}
\usepackage{subcaption} 
\usepackage{wrapfig,lipsum}
\usepackage{balance}

\newcommand{\ri}{\text{(\rom 1)~}}
\newcommand{\rii}{\text{(\rom 2)~}}
\newcommand{\riii}{\text{(\rom 3)~}}
\newcommand{\riv}{\text{(\rom 4)~}}

\newcommand{\model}{{{\tt DCCL}}}

\usepackage{xr-hyper}
\makeatletter
\newcommand*{\addFileDependency}[1]{
  \typeout{(#1)}
  \@addtofilelist{#1}
  \IfFileExists{#1}{}{\typeout{No file #1.}}
}
\makeatother

\newcommand*{\myexternaldocument}[1]{
    \externaldocument{#1}
    \addFileDependency{#1.tex}
    \addFileDependency{#1.aux}
}
\myexternaldocument{CL4DG_supp}

\usepackage{paperpkg}
\usepackage{soul}
\usepackage{booktabs}

\newcommand{\yc}[1]{\textcolor{teal}{#1}}
\newcommand{\tx}[1]{\textcolor{black}{#1}}

\usepackage[capitalize]{cleveref}
\crefname{section}{Sec.}{Secs.}
\Crefname{section}{Section}{Sections}
\Crefname{table}{Table}{Tables}
\crefname{table}{Tab.}{Tabs.}

\AtBeginDocument{%
  \providecommand\BibTeX{{%
    \normalfont B\kern-0.5em{\scshape i\kern-0.25em b}\kern-0.8em\TeX}}}

\copyrightyear{2025}
\acmYear{2025}
\setcopyright{rightsretained}
\acmConference[KDD '25] {Proceedings of the 31st ACM SIGKDD Conference on Knowledge Discovery and Data Mining V.1}{August 3--7, 2025}{Toronto, ON, Canada}
\acmBooktitle{Proceedings of the 31st ACM SIGKDD Conference on Knowledge Discovery and Data Mining V.1 (KDD '25), August 3--7, 2025, Toronto, ON, Canada}
\acmDOI{10.1145/3690624.3709280}
\acmISBN{979-8-4007-1245-6/25/08}

\makeatletter
\newcounter{pagecount}
\newcommand{\limitpages}[1]{
    \setcounter{pagecount}{0}%
    \gdef\maxpages{#1}%
    \ifx\latex@outputpage\@undefined\relax%
        \global\let\latex@outputpage\@outputpage%
    \fi%
    \gdef\@outputpage{%
        \addtocounter{pagecount}{1}%
        \ifnum\value{pagecount}>\maxpages\relax%
        \else%
            \latex@outputpage%
        \fi%
    }%
}
\makeatother

\begin{document}

\title{Connecting Domains and Contrasting Samples: A Ladder for Domain Generalization
}



\author{Tianxin Wei}
\authornote{Tianxin and Yifan contributed equally to this work.}
\affiliation{  \institution{
UIUC}
  \city{Champaign}
  \state{IL}
  \country{USA}}
\email{twei10@illinois.edu}

\author{Yifan Chen}
\authornotemark[1]
\affiliation{
  \institution{
HKBU}
  \city{Kowloon}
  \country{HK}
}
\email{yifanc@hkbu.edu.hk}

\author{Xinrui He}
\affiliation{
  \institution{
UIUC}
  \city{Champaign}
  \state{IL}
  \country{USA}
}
\email{xhe33@illinois.edu}

\author{Wenxuan Bao}
\affiliation{
  \institution{
UIUC}
  \city{Champaign}
  \state{IL}
  \country{USA}
}
\email{wbao4@illinois.edu}

\author{Jingrui He}
\affiliation{  \institution{
UIUC}
  \city{Champaign}
  \state{IL}
  \country{USA}}
\email{jingrui@illinois.edu}


\begin{abstract}
Distribution shifts between training and testing samples frequently occur in practice and impede model generalization performance. 
This crucial challenge thereby motivates studies on domain generalization (DG), which aim to predict the label on unseen target domain data by solely using data from source domains. 
It is intuitive to conceive the class-separated representations learned in contrastive learning (CL) are able to improve DG,
while the reality is quite the opposite: users observe directly applying CL deteriorates the performance.
We analyze the phenomenon with the insights from CL theory and discover lack of \emph{intra-class connectivity} in the DG setting causes the deficiency.
We thus propose a new paradigm, domain-connecting contrastive learning (\model), to enhance the conceptual connectivity across domains and obtain generalizable representations for DG. 
On the data side, more aggressive data augmentation and cross-domain positive samples are introduced to improve \textcolor{black}{intra-class connectivity}. 
On the model side, to better embed the unseen test domains, we propose model anchoring to exploit the \textcolor{black}{intra-class connectivity} in pre-trained representations and complement the anchoring with generative transformation loss.
Extensive experiments on five standard DG benchmarks are performed. The results verify that \model~outperforms state-of-the-art baselines even without domain supervision.
The detailed model implementation and the code are provided through \url{https://github.com/weitianxin/DCCL}


\end{abstract}





\maketitle



\section{Introduction}
\label{sec:intro}
Modern machine learning has achieved great progress in various applications, such as computer visual \citep{he2016deep,tan2020efficientdet,cheng2021per,sun2019deep,weirobust,zhang-etal-2023-vibe}, and natural language processing \citep{NIPS2017_3f5ee243,devlin2019bert,chen2021skyformer,raffel2023exploring,wei2023ntk,he2024llm}. 
Despite the immense success, existing approaches 
typically assume that training and testing data are independently sampled from the identical distribution.
However, in real-world scenarios, 
this assumption rarely holds. 
In image recognition, for example, distribution shifts
w.r.t.\ geographic locations \citep{beery2018recognition} and image background \citep{fang2013unbiased} frequently occur and impede models' generalization performance.


\begin{figure*}[t]
     \centering
     \begin{subfigure}[t]{0.24\textwidth}
         \centering
         \includegraphics[width=0.95\textwidth]{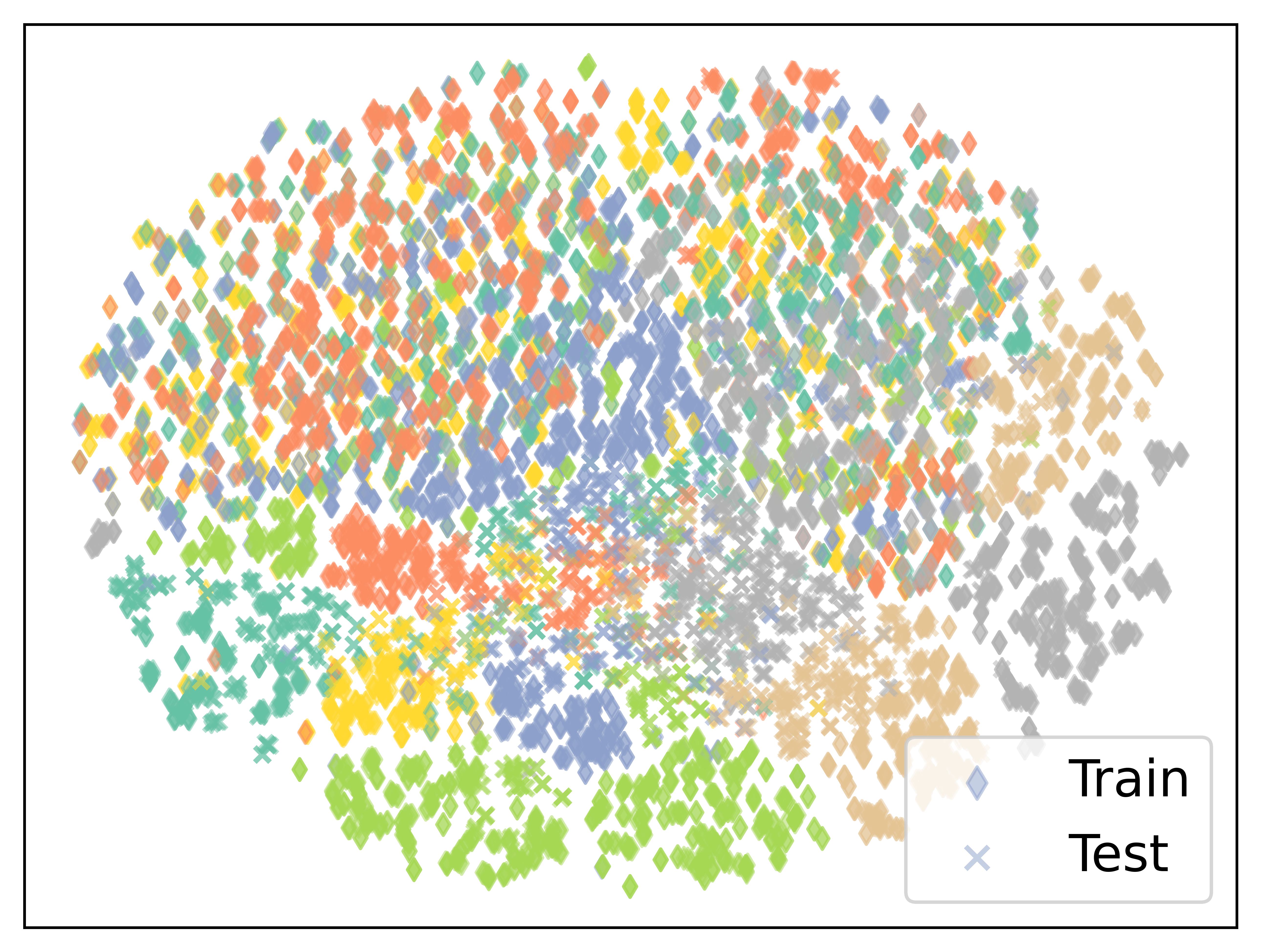}
         \caption{Pre-trained.}
         \label{fig:pretrained_domain}
     \end{subfigure}
     \begin{subfigure}[t]{0.24\textwidth}
         \centering
         \includegraphics[width=0.95\textwidth]{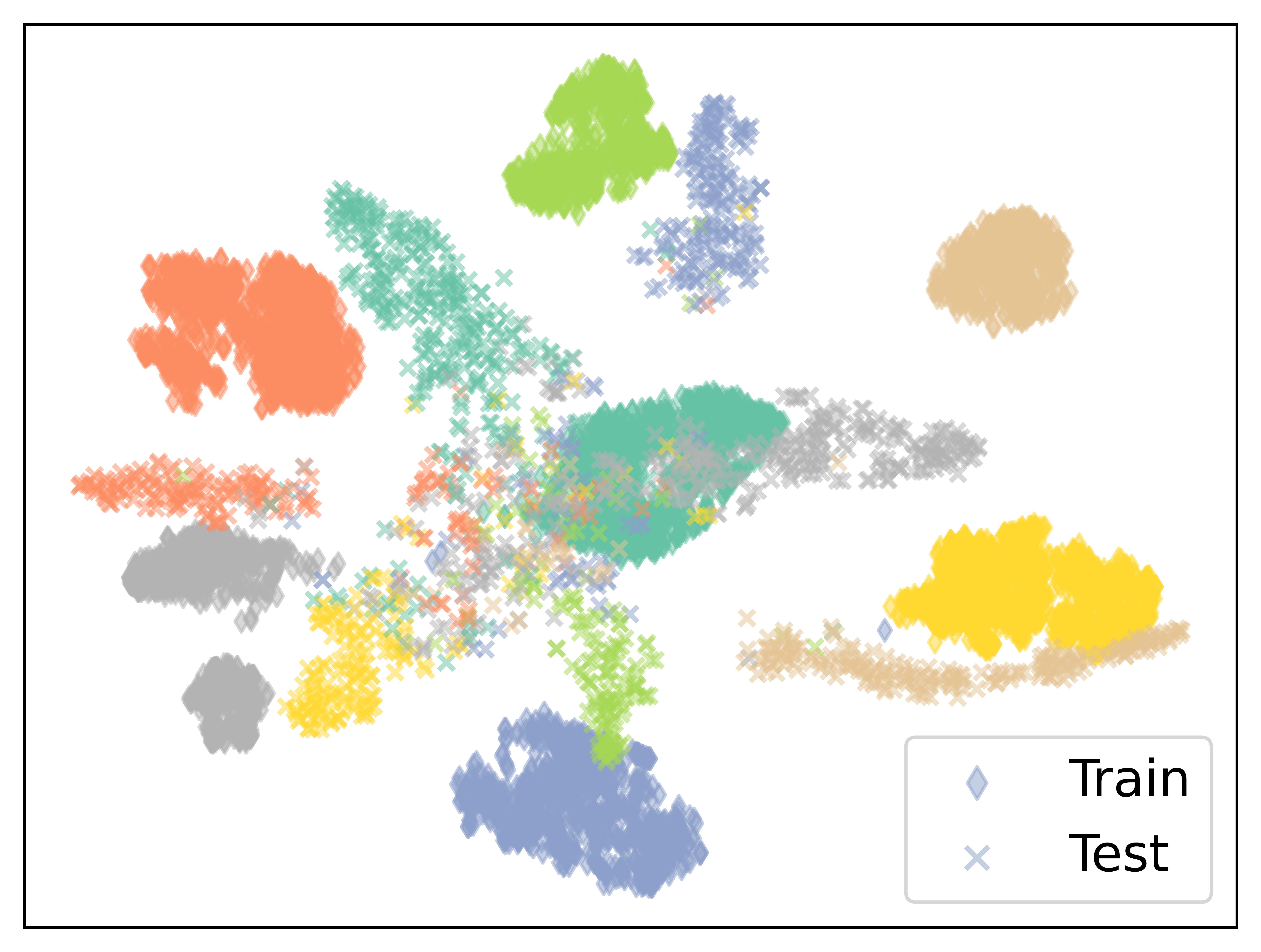}
         \caption{ERM.}
         \label{fig:erm_domain}
     \end{subfigure}
     \begin{subfigure}[t]{0.24\textwidth}
         \centering
         \includegraphics[width=0.95\textwidth]{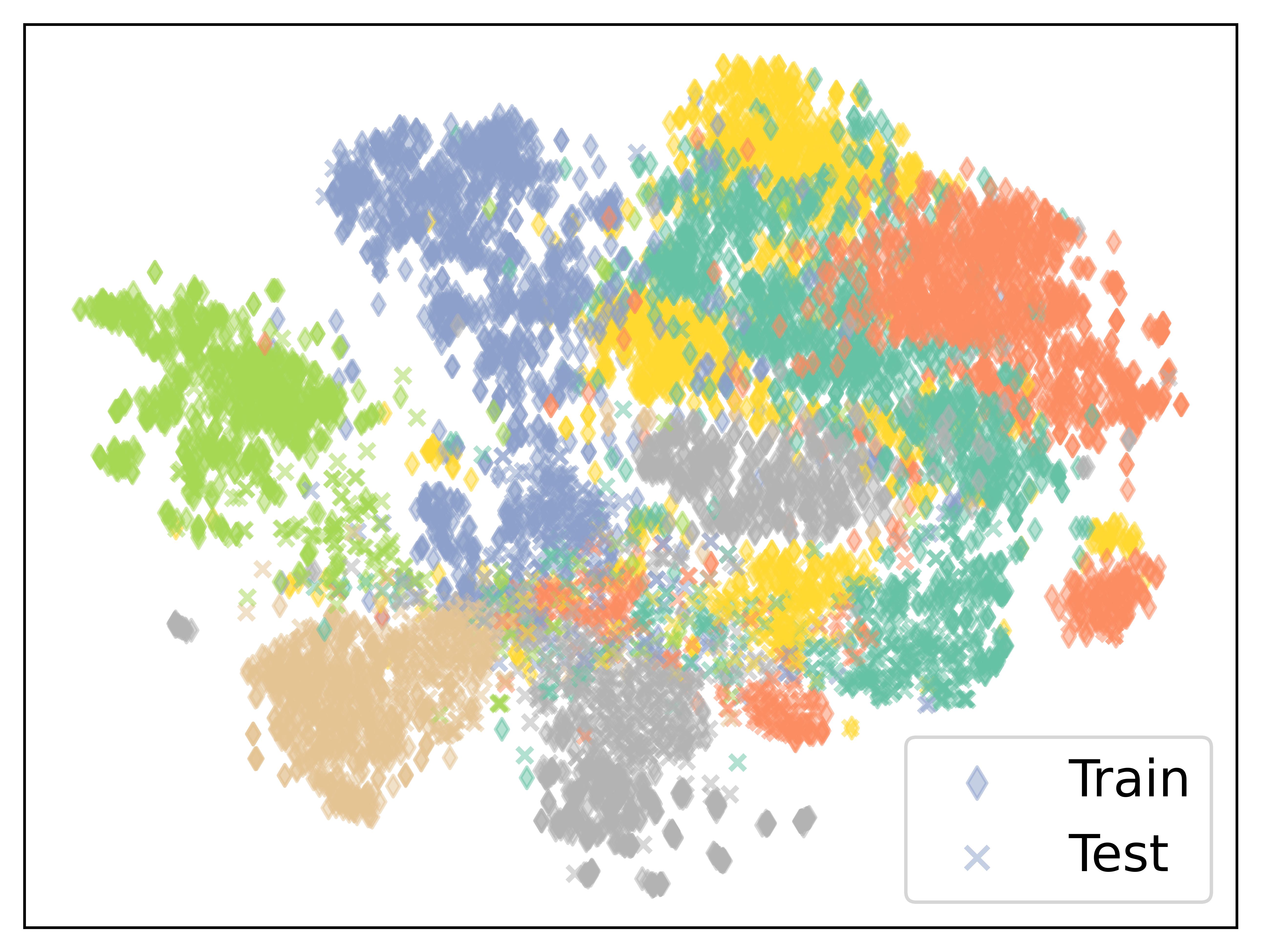}
         \caption{SCL.}
         \label{fig:scl_domain}
     \end{subfigure}
     \begin{subfigure}[t]{0.24\textwidth}
         \centering
         \includegraphics[width=0.95\textwidth]{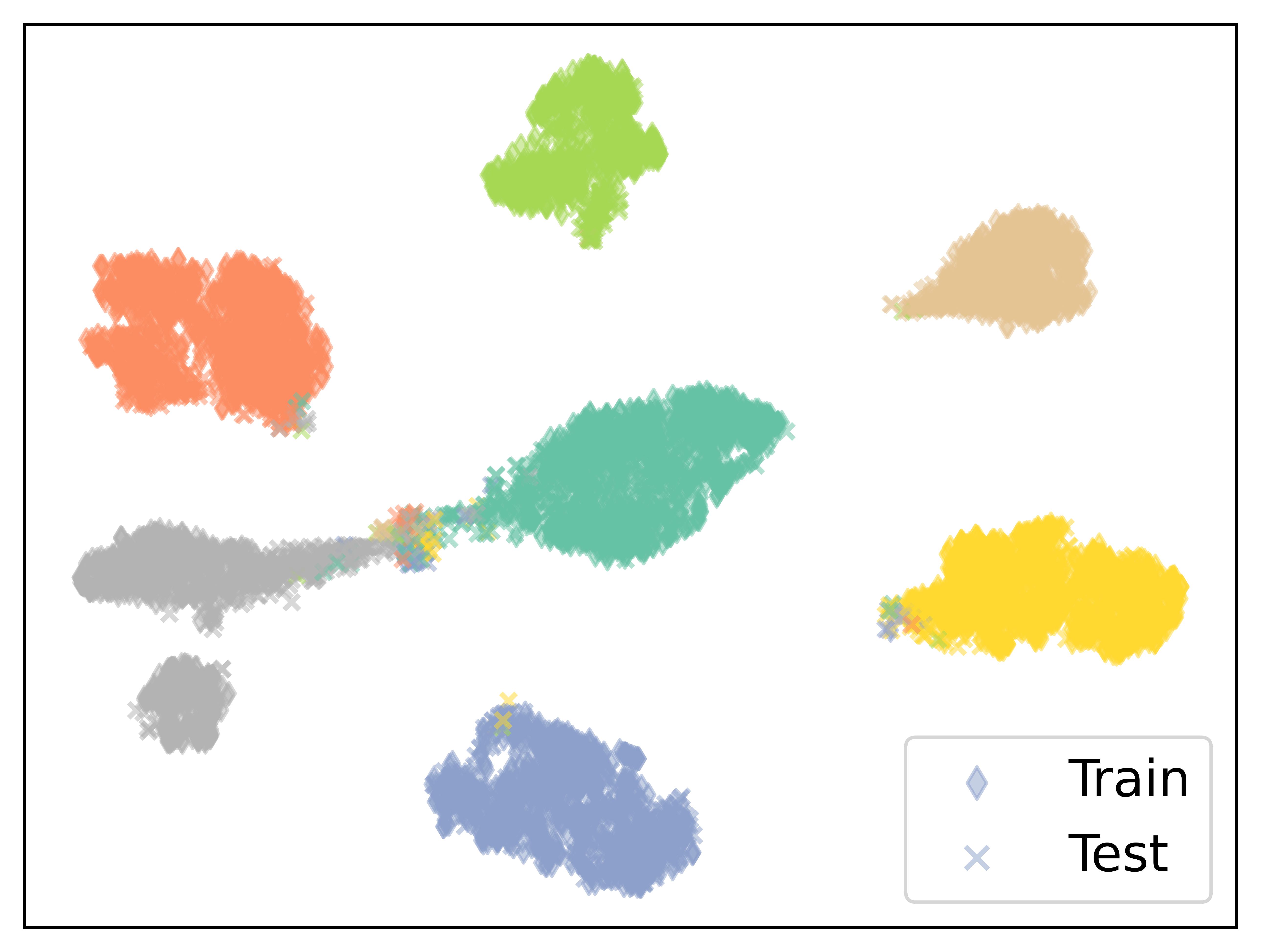}
         \caption{\model.}
         \label{fig:dccl_domain}
     \end{subfigure}
        \caption{t-SNE visualization of the representations across both training and testing domains, output by Pre-trained, ERM, SCL and our \model~respectively. 
        Same-class points share colors, while marker types differentiate training and testing domains. (Please zoom in for better viewing.)
        We visualize the embedding on PACS dataset where the source domains are [Photo], [Sketch], and [Cartoon]; the target domain is [Art]. Note that when mapped by the pre-trained model, intra-class samples from both the training and testing domains appear scattered but indeed well-connected. SCL will lead to a degradation in the embedding quality. Our proposed \model, on the other hand, effectively clusters the intra-class samples.}
        \label{fig:cross_domain}
        \vspace{-0.2cm}
\end{figure*}

Accordingly, domain generalization (DG)~\citep{gulrajani2020search} is studied to enhance the transferability of deep learning models.
A natural idea for DG is to learn invariant representations for same-class samples across a variety of seen domains so as to benefit the classification of unobserved testing domain samples.
As a powerful representation learning technique, contrastive learning (CL) \citep{chen2020simple} aims to obtain class-separated representations and has the potential for DG \citep{kim2021selfreg}.
In this paper, however, we have observed the limitation of the widely deployed self-contrastive learning (SCL), which aligns the augmentation of the same input. Although SCL has demonstrated success in unsupervised pre-training tasks \cite{chen2020simple,he2020momentum,grill2020bootstrap}, 
it does not naturally fit the domain generalization setting: SCL implicitly assumes the capability to sample instances from the whole data distribution, which does not fit the practical domain generalization scenario where models are fine-tuned using data from specific partial domains. Consequently, SCL struggles to acquire generalizable representations in this context.


To bridge this gap, we propose domain-connecting contrastive learning (\model) to pursue transferable representations in DG, whose core insight comes from a recent novel understanding attributing the success of CL to the intra-class representation connectivity~\citep{wang2022chaos}.
Specifically, we first suggest two direct approaches to improve \emph{\textcolor{black}{intra-class connectivity}} (to be fully explained at the beginning of \Cref{section:problem}) within CL models: applying more aggressive data augmentation and expanding the scope of positive samples from self-augmented outputs to the augmentation of same-class samples across domains. 
The aforementioned approaches aid in establishing connections among existing domains. 

The module above focuses on enhancing intra-class connectivity from the data perspective. However, the embeddings of the unseen testing domains and the ones of the training domains in the same class may still be separated. 
To address this issue, we make and utilize an observation that the pre-trained models from the large database, unlike the learned maps of Empirical Risk Minimization (ERM), indeed \textbf{possess the desired intra-class connectivity}: the intra-class samples of the training domains and the testing domains are scattered but well-connected, as demonstrated in Figure \ref{fig:pretrained_domain} and Section \ref{subsection:pre-train}. This encouraging observation motivates us to anchor learned maps to the pre-trained model by broadening the augmentation strategies in CL.

Furthermore, to close the gap in the representations of pre-trained and fine-tuned models, we propose to complement contrastive learning with the generative transformation loss for enriched supervised signals.
As a visual illustration, Figure~\ref{fig:cross_domain} demonstrates the embeddings learned by regular ERM and by the proposed \model.
ERM embeds the data in a more scattered distribution, and many samples in the central region cannot be distinguished;
on the other hand, \model~well clusters inter-class samples regardless of the domains. 
It verifies the effectiveness of our proposed \model~on connecting domains. Our contributions are summarized as follows:
\begin{itemize}[leftmargin=*]
\item We analyze the failure of self-contrastive learning on DG and propose two effective strategies to improve \textcolor{black}{intra-class connectivity} within CL models.
\item We propose to anchor learned maps to pre-trained models that possess the desired connectivity of training and testing domains.
We further propose generative transformation loss to complement the alignment between learned maps and pre-trained models.
\item We conduct extensive experiments on five real-world DG benchmarks with various settings, demonstrating the effectiveness and rationality of \model.
\end{itemize}

The rest of the paper is organized as follows. We introduce the problem formulation and preliminaries in Section \ref{section:problem}, present our proposed \model~in Section \ref{section:method}, show the experimental results in Section \ref{section:experiments}, discuss the related work in Section \ref{section:related work}, and conclude in Section \ref{section:conclusion}.

\section{Preliminaries}
\label{section:problem}

We first illustrate the core concept of the paper, \emph{\textcolor{black}{intra-class connectivity}}. 
It refers to the intra-class data connectivity across different domains and resembles the connectivity in CL theory~\citep{wang2022chaos}, which depicts the preference that samples should not be isolated from other intra-class data of the same class~\footnote{
An intuitive graph-based measure to assess the \textcolor{black}{intra-class connectivity} of a given model is discussed in \Cref{app:connectivity}
}.
In the remainder of this section, we introduce the problem formulation and necessary preliminaries for contrastive learning. A thorough review of related work on domain generalization and contrastive learning are deferred to \Cref{section:related work}.

\subsection{Data in the Domain Generalization Setting}


Given $N$ observations (from $M$ domains), $\mathbf{X} = \{x_1, \dots, x_N\} \subseteq \m X$ is the collection of input designs, $\mathbf{Y}=\{y_1, ..., y_N\} \subseteq \m Y$ represents the prediction targets,
and the whole dataset $D_s$ is denoted as $\{(x_i^m, y_i^m)_{i=1}^{N^m}\}_{m=1}^M$, where $N^m$ is the number of samples (naturally, $\sum_{m=1}^M N^m = N$) in domain $d^m$ and $x_i$ is re-indexed as $x_i^m$.

\subsection{Model Optimization with Contrastive Learning}
Contrastive Learning (CL) enforces the closeness of augmentation from the same input, compared to other inputs in the representation space. 
The main components of CL, as summarized in \cite{chen2020simple,he2020momentum}, include:
\ri data augmentation for contrastive views, \rii a representation map $f$ as the data encoder: $\mathcal{X}\rightarrow \mathbb{R}^{d}$, \riii projection head $h(\cdot)$ for expressive representation, and \riv the contrastive loss for optimization. 
Given an instance from $\mathbf{X}$, we draw a positive pair $x, x^{+}$ by applying a random data augmentation $a \sim \m A$, where $\m A$ is the pre-specified distribution of random data augmentation maps. 
As a contrastive concept to positive samples, a negative pool 
$\mathcal{N}_x$ is the set of augmented samples randomly drawn from the whole dataset~$\mathbf{X}$. 
To ease the construction of the CL loss, 
we denote $p(x)$ as the distribution of $x$, $p\paren{x, x^+}$ as the corresponding joint distribution of the positive pairs, and $p_n(x_i^-)$ (``n'' is shorthand for ``negative'') as the distribution for the negative sample $x_i^- \in \m N_x$, which are all independent and identically distributed (i.i.d.). 
Let $z$ denote the normalized output of input feature $x$ through $f_h \defeq \paren{h \circ f}(\cdot)$. Consequently, $z^{+}=f_h(x^{+})$ is the embedding for the positive sample of $z=f_h(x)$ , and ${z_i}^{-}=f_h(x^{-}_i)$ represents the embedding of the samples in the negative pool $\mathcal{N}_x$. 

The most common form of the CL loss ($\mathcal{L}_{\text{CL}}$) adapts the earlier InfoNCE loss~\citep{oord2018representation}, formulated as:
\begin{equation}
    \mathcal{L}_{\text {CL }} = \underset{\substack{p({x, x^+}) \\ \set{p_n(x_i^-)}_{i=1}^{\abs{\m N_x}}}}{\Expect} \left[-\log 
    \frac{\exp \left(z \cdot z^{+} / \tau \right)}
    {\underset{i \in \brkt{\abs{\m N_x}}}{\sum} \exp \left(z \cdot z_{i}^{-} / \tau\right)}\right]
\label{eqn:cl loss}
\end{equation}
where $\tau > 0$ is the temperature parameter. The CL loss is typically used in the unsupervised \citep{chen2020simple,he2020momentum,grill2020bootstrap} or supervised \citep{khosla2020supervised} \textbf{pre-training} setting. To adapt it to \textbf{domain generalization} \citep{yao2022pcl,chen2022compound,kim2021selfreg}, the full model is also required to learn from supervised signals. 
Thus, it is intuitive to combine the CL loss with the empirical risk minimization (ERM) loss $\mathcal{L}_{\text{ERM}}$ as the following objective:
\begin{equation}
    \mathcal{L}=\mathcal{L}_{\text{ERM}}+\lambda \mathcal{L}_{\text {CL }}
\end{equation}
where $\lambda$ is the regularization hyper-parameter during training. 
In practice, $\mathcal{L}_{\text{ERM}}$ is usually chosen as the softmax cross entropy loss to classify the output embedding $z$. We follow the classical setting \citep{kim2021selfreg} in this paper, which includes both classification loss and self-supervised regularization loss.

\begin{figure}[tbp]
\centering
  \includegraphics[width=\columnwidth]{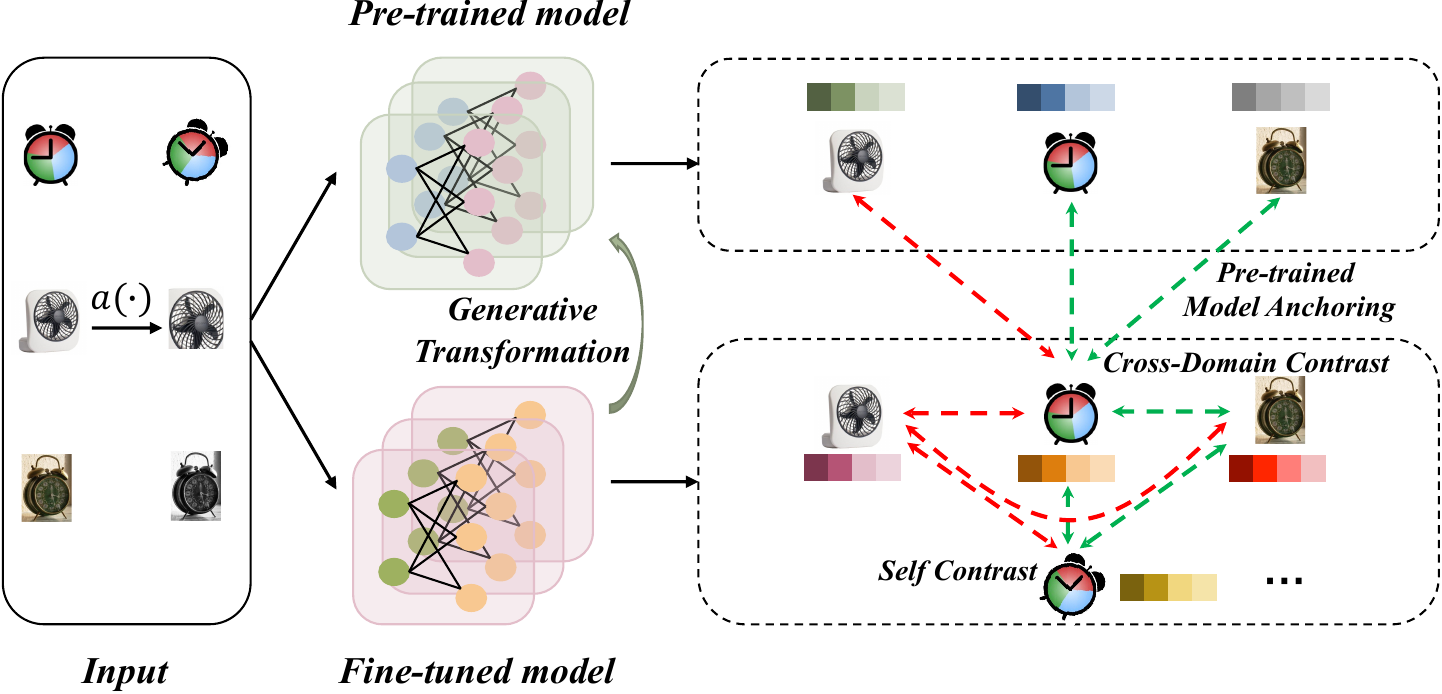}
  \caption{The overall framework of \model. The green dotted arrows indicate the two representations form a positive pair and the red ones connect the negative pairs. $a(\cdot)$ is an aggressive augmentation operation. Two key parts in \model~are \ri cross-domain data contrast to bridge the intra-class samples across domains; \rii pre-trained model anchoring, completed with generative transformation to harness the intra-class connectivity inherent in the pre-trained representation.}
\label{fig:framework}
\vspace{-0.4cm}
\end{figure}

\section{Proposed Methodology}
\label{section:method}

In this section, we present the details of \model,
which learns robust representations for tackling distribution shifts across domains. 
We first comment on the failure of directly applying self-contrastive learning to DG in \Cref{subsection:failure}. 
Followed by the implications from learning theory in Section~\ref{subsection:implication}, we propose two complementary strategies to improve intra-class data connectivity in Section~\ref{subsection:strategy} to initialize our domain-connecting CL. 
Then in Section~\ref{subsection:pre-train}, we introduce pre-trained model anchoring to further utilize the \textcolor{black}{intra-class connectivity} of the representation output by the pre-trained model. 
A generative transformation module is designed to assist the anchoring and help encode the essential information in the pre-trained representation. The overall framework of \model~is shown in Figure~\ref{fig:framework}, which integrates data and model information for generalization.


\begin{figure*}[t]
     \centering
     \begin{subfigure}[t]{0.45\textwidth}
         \centering
         \includegraphics[width=0.93\textwidth]{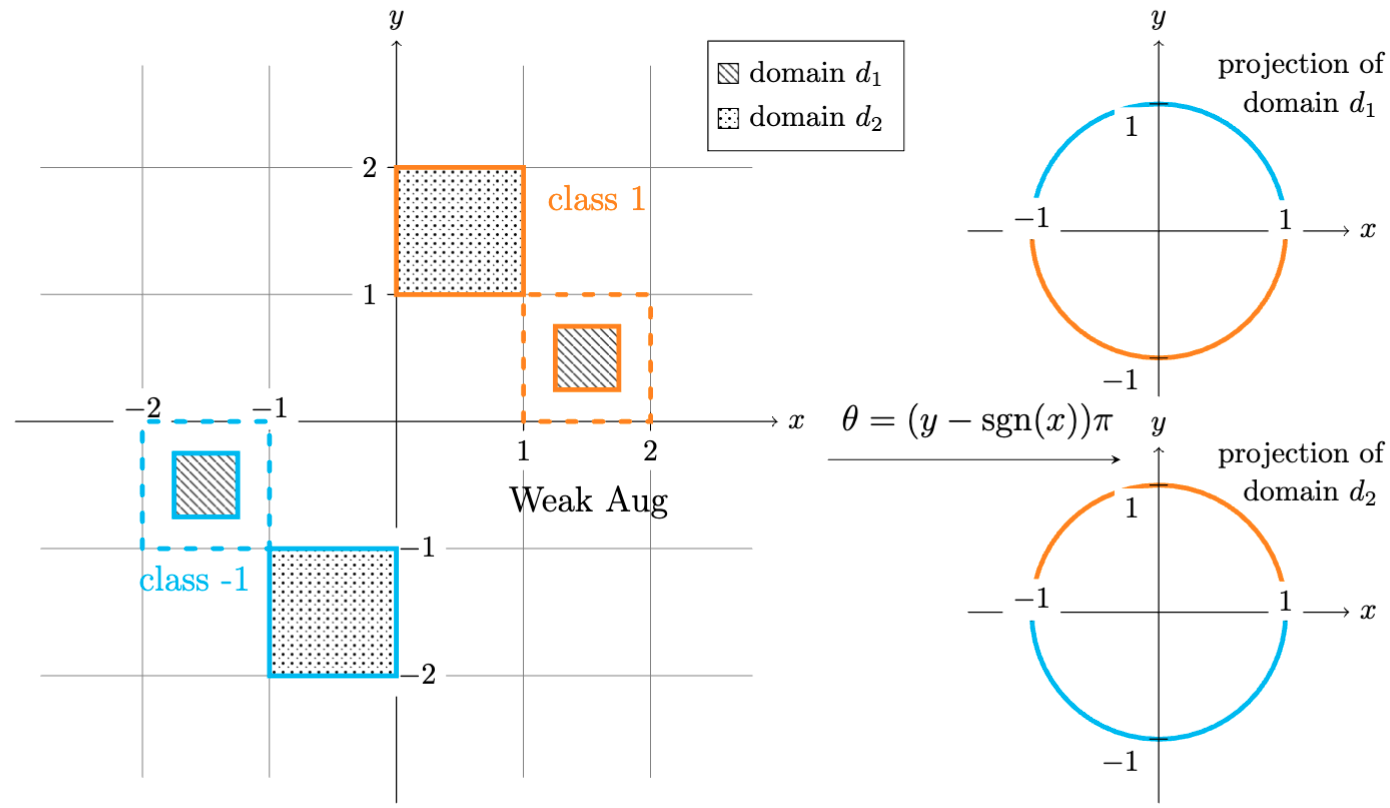}
         \caption{Weak Augmentation.}
         \label{fig:weak}
     \end{subfigure}
     \begin{subfigure}[t]{0.45\textwidth}
         \centering
         \includegraphics[width=0.93\textwidth]{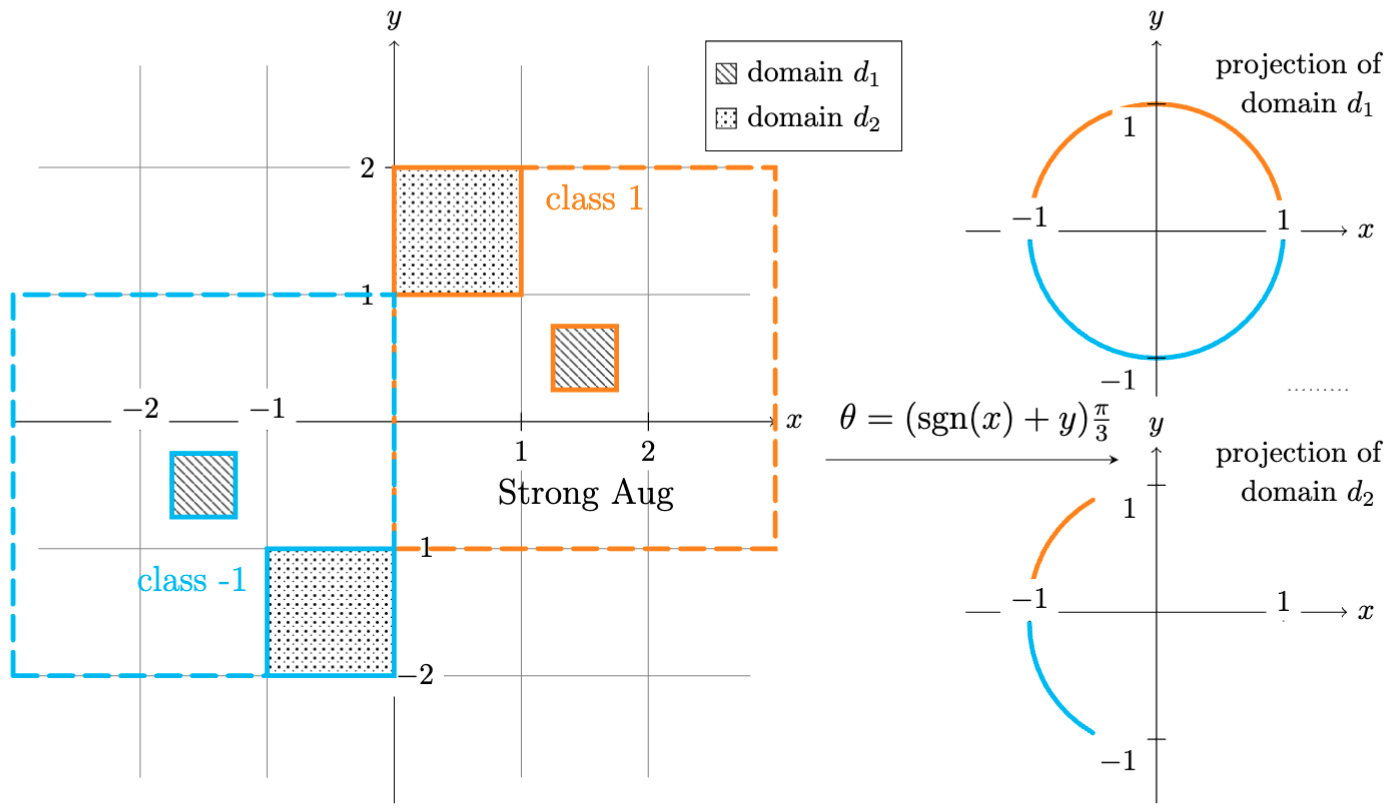}
         \caption{Aggressive Augmentation.}
         \label{fig:strong}
     \end{subfigure}
        \caption{Illustration for the toy example of self-contrastive learning (SCL). Spots and slashes are filled in to represent different domains; orange and black rectangles respectively denote classes 1 and 2. The mapping function $\varphi \circ \theta$ learned on domain $d_1$ can perfectly classify the samples, and the mapping attains perfect alignment and uniformity (the objective of SCL). When trained with weak augmentation and applied to a new domain $d_2$, the classifier completely fails (0\% acc). With aggressive augmentation, the intra-class samples of different domains are connected and we obtain transferable representations (100\% acc).}
        \label{figure:example}
        \vspace{-0.4cm}
\end{figure*}

\subsection{Motivation: Failure of Self-contrastive Learning in Domain Generalization}
\label{subsection:failure}

Self-contrastive learning, which aligns the augmentation views of the same input, has achieved impressive performance in unsupervised pre-training tasks \citep{chen2020simple,he2020momentum,grill2020bootstrap}. However, it does not naturally fit the domain generalization setting since it assumes the ability to sample $x$ from the whole data distribution:
in the training stage of domain generalization, we instead are only able to access partial domains.
This mismatch can lead to suboptimal performance in DG if the users mechanically adopt the classical CL loss. 

We provide a linearly separable toy example in Figure~\ref{figure:example} to show the deficiency of SCL. In particular, even attaining the optimal CL loss (\ref{eqn:cl loss}) cannot guarantee good DG performance, 
where only partial domains are involved in training. 
We detail the coined data distribution as follows.

\begin{example}[SCL does not help domain generalization.]
\label{eg:cl loss failure}
Let the label collection $\m Y$ be $\set{-1, 1}$ and the portions of two classes be both $0.5$.
Assume there are two domains $d_1$ and $d_2$:
if a sample $X = (X_1, X_2) \in \mb R^2$ with label $Y$ is from domain $d_1$, its conditional distribution will be specified as
\begin{align*}
\begin{cases}
      X_1 \sim \mathrm{Unif}\paren{1.25, 1.75} Y, \\
      X_2 \sim \mathrm{Unif}\paren{0.25, 0.75} Y, \\
      X_1 \indep X_2 \mid Y;
\end{cases}
\end{align*}
In domain $d_2$ the distribution of $X_1, X_2$ can be analogously represented.
Considering only domain $d_1$ is involved in training, we construct a map $\varphi \paren{\theta(x)} \defeq \paren{\cos\paren{\theta}, \sin\paren{\theta}}$ with $\theta(x) = \paren{x_1 - \mathrm{sgn}(y)} \pi$ for the weak augmentation setting and $\theta(x) = \paren{\mathrm{sgn}(x_1)+y} \frac{\pi}{3}$ for the aggressive augmentation setting. 
The map $f_h = \varphi \circ \theta$ attains perfect alignment of intra-class samples and maximal uniformity (representations of the augmented samples are uniformly distributed on the corresponding circle arcs) on the $1$-sphere $\mb S^1 \defeq \set{x \in \mathbb{R}^{2}:\|x\|_2=1}$. Based on the derivation in \cite{wang2020understanding}, $f_h$ will minimize the CL loss~(\ref{eqn:cl loss}).
\end{example}

Figure \ref{figure:example} illustrates the example, where slashes and spots are used to represent domains $d_1$ and $d_2$; orange and blue rectangles respectively denote classes 1 and -1. For ease of analysis, we specifically consider the case that only domain $d_1$ is involved in training. Note that adding more domains does not affect the conclusion of our analysis. In Figure \ref{fig:weak}, We can observe that when applying weak augmentation, the new representations for domain $d_2$ do not reflect the class information and even have the opposite signs as domain $d_1$. On the other hand, in Figure \ref{fig:strong}, with aggressive augmentation, the intra-class samples of different domains are connected. In this case, the optimal representations learned on domain $d_1$ can also reflect the accurate class information of testing domain $d_2$.

We can conclude that the usage of classical SCL with weak augmentation does not necessarily lead to good DG performance; empirical verification is provided in Section~\ref{sec:exp_case} as well. A similar limitation is observed in invariance-based DG methods~\citep{shui2022benefits}. The key to the problem lies in improving the intra-class connectivity (achieved by aggressive augmentation in this example) across domains.

\subsection{Implications from Contrastive Learning Theory}
\label{subsection:implication}

Building on these observations, we delve deeper into understanding these limitations. Specifically, we demonstrate that intra-class connectivity is crucial for reducing the intra-class representation variance $\mathrm{Var}(f_h(x) | y)$, as outlined in Proposition \ref{prop:invariance} and further analyzed in Appendix \ref{app:analysis} due to space limitations. Reducing this variance enhances domain generalization by promoting stable feature representations that are less influenced by domain-specific variations. This theoretical framework of connectivity motivates us to re-examine the failures of SCL discussed in the previous subsection, focusing on the connectivity perspective to uncover potential solutions and improvements.




With regard to domain generalization, if all intra-class samples can be clustered together across domains and the intra-class variance shrinks to zero in CL, we then automatically obtain the generalizable representations.
We observe that SCL in the previous example fails to obtain \textbf{intra-class connectivity} due to insufficient data augmentation and domain-separated (rather than class-separated) representations, which ultimately cause poor generalization performance. 
We thus propose two approaches to improve intra-class connectivity: \ri applying more aggressive data augmentation and \rii expanding the scope of positive samples, from solely self-augmented output $a(x)$ to the augmentation of intra-class samples across domains.
In applying CL, proper data augmentation can help ``connect'' two different samples $x_i, x_j$ within the same class, which technically means there exists a pair of augmentation maps $a_i, a_j$ so that $a_i(x_i), a_j(x_j)$ are close to each other.
Consequently, in optimizing the CL loss (\ref{eqn:cl loss}) the representations $f_h(x_i), f_h(x_j)$ will be pushed close since
\[
f_h(x_i) \approx f_h\paren{a_i(x_i)} \approx f_h\paren{a_j(x_j)} \approx f_h(x_j).
\]
In other words, as a ladder, $a_i(x_i)$ and $a_j(x_j)$ connect the two samples $x_i, x_j$, and analogously all the samples within the same class can be connected by proper data augmentation. Similarly, expanding the scope of positive samples can help connect the samples from different domains but same classes, and thus enhance the intra-class connectivity. 
CL later on pushes their learned representations to cluster thanks to the CL loss. 

We remark IRM \citep{arjovsky2019invariant} proposed a similar idea of leveraging the intra-class sample similarities,
while the CL theory removes the assumption in IRM that the marginal distribution of sample $x$ on source domains should be the same on target domains,
and thus is theoretically more applicable to DG.



\subsection{More Aggressive Data Augmentation and Cross-domain Positive Samples}
\label{subsection:strategy}

Inspired by the analysis above, we propose two direct approaches to improve \textcolor{black}{intra-class connectivity}: \ri applying more aggressive data augmentation and \rii expanding the scope of positive samples, from solely self-augmented output $a(x)$ to the augmentation of intra-class samples across domains.

For the first approach, despite the fact that data augmentation in DG (e.g., horizontal flipping) has already been a standard regularization technique \citep{gulrajani2020search,cha2021swad,wang2022generalizing},
the choice of data augmentation, we emphasize, matters for CL in the DG setting.
We naturally need a larger augmentation distribution $\m A$ to connect $a_i(x_i)$ and $a_j(x_j)$ since $x_i, x_j$ can be drawn from different domains.
The effect of data augmentation intensity is evaluated through the ablation studies in Section~\ref{sec:exp ablation}.

Motivated by supervised CL~\citep{khosla2020supervised,gunel2020supervised,cui2021parametric}, we further introduce \textbf{cross-domain} positive pairs into CL to bridge the intra-class samples scattered in different domains. 
Specifically, we not only consider the correlated views of the same data sample as positive pairs but also the augmented instances from other intra-class samples across domains.
The positive sample $x^+$ will now be conditionally independent of $x$,
and the positive pairs have the same conditional distribution $p^{(1)}(x^+|y) = p(x|y)$
\footnote{Unlike the classical setting in self-supervised CL, in DG we can access the label $y$ in training.} (the specific distribution of the positive sample $x^+$ in this subsection will be denoted with a superscript $(1)$);
in other words, $x^+$ can now be the augmentation view of a random sample within the same class $y$ of $x$.
With the joint distribution of $x, x^+$ denoted as 
$p^{(1)}\paren{x, x^+}=\int_y p^{(1)}(x^+|y) p(x|y) p(y) \dd y$,
the primal domain-connecting contrastive learning (\model) objective $\mathcal{L}_{\model}^{(0)}$ can be formulated as:
\begin{equation}
     \mathcal{L}_{\model}^{(0)} = \underset{\substack{p^{(1)}(x, x^+)  \\ \set{p_n\paren{x_i^-}}_{i=1}^{\abs{\m N_x}}}}{\mathbb{E}}
     \left[-\log \frac{\exp \left(z \cdot z^{+} / \tau\right)}{\underset{i \in \brkt{\abs{\m N_x}}}{\sum} \exp \left(z \cdot z_{i}^{-} / \tau\right)}\right].
\label{eqn:dccl loss 1}
\end{equation}

Unlike supervised CL, which forms positive pairs from different views within the same domain, our method incorporates intra-class samples across domains, effectively improving intra-class connectivity from a data perspective. The term, $-\log \exp \left(z \cdot z^{+} / \tau\right)$, corresponding to alignment in loss~(\ref{eqn:dccl loss 1}), can now push the intra-class samples from different domains together.



\subsection{Anchoring Learned Maps to Pre-trained Model}
\label{subsection:pre-train}

Up to now, we have not addressed the core challenge in DG---lack of access to the testing domains in training: 
CL is originally designed for the self-supervised scenario where a huge amount and wide range of data is fed to the models. 
However, in the context of domain generalization, the model is just fine-tuned on limited data within partial domains.
Consequently, the mechanism of CL can only contribute to the clustering of representations in the seen domains, while the embeddings of the unseen testing domains and the ones of the training domains in the same class may still be separated.

Interestingly, the \textcolor{black}{intra-class connectivity} for representations, the desired property in CL, seems to exist at the beginning of the fine-tuning.
We observe the phenomenon when visualizing the representations obtained from the pre-trained model using t-SNE~\citep{van2008visualizing} in Figure~\ref{fig:pretrained_domain},
which thereby motivates our design in this subsection. 
We find that mapped by the initial pre-trained model ResNet-50, intra-class samples of the training domains and the testing domains are scattered while well-connected.

We attribute the phenomenon to the effective representations returned by pre-trained models, which reasonably model the pairwise interactions among samples and thus draw target domains closer to source domains. 
To verify the effectiveness of the representations, we design a quantitative \textbf{metric to evaluate} whether the pre-trained space is ``well-connected'', by turning to the concept of ``connectivity'' in graphs. Details can be found in \Cref{app:connectivity}.

As for the model design, the phenomenon motivates us to better utilize the pre-trained model $f_{\mathrm{pre}}$ for stronger \textcolor{black}{intra-class connectivity} in the mapped representations obtained from $f$.
We propose to make use of pre-trained models as data augmentation in a disguised form: data augmentation works on the raw data while we can further ``augment'' the representation $x$ via the model $f_{\mathrm{pre}}$.

\begin{figure}[t]
    \centering
    \includegraphics[width=0.9\columnwidth]{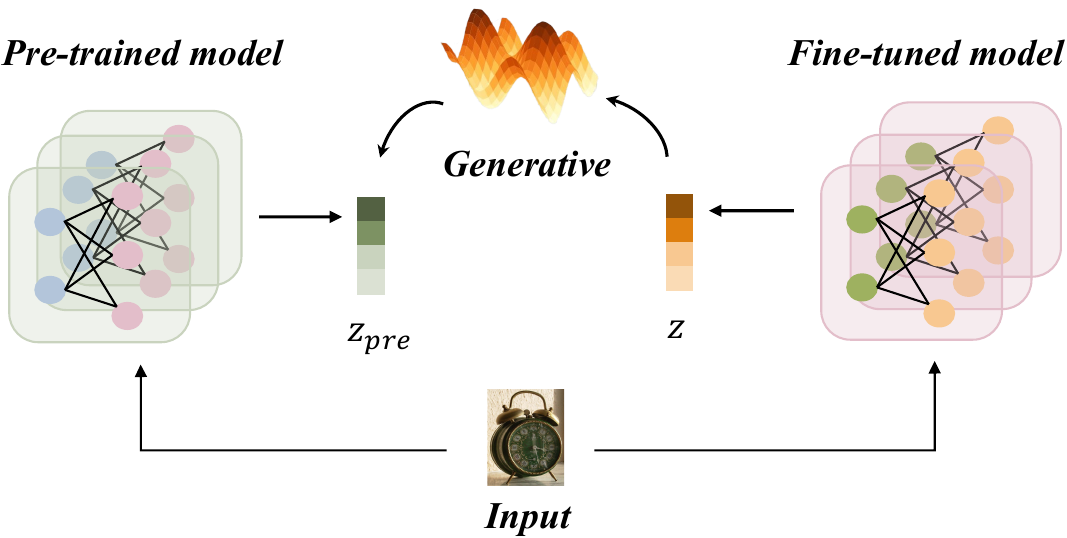}
    \caption{An overview of the generative transformation module in \model. Two representations $z_{pre}$ and $z$ of the same image are generated via the pre-trained and the fine-tuned model respectively. The variational reconstruction is conducted to encode essential within-sample information.}
\label{fig:gen}
\vspace{-0.3cm}
\end{figure}

In mathematical language, in additional to the augmented sample $x^+$ defined in the last subsection, we further incorporate the pre-trained embedding $z_\mathrm{pre}  = h \circ f_{\mathrm{pre}}(x)$ into the definition of feasible positive embeddings $z^{(2), +}$, which expands the scope of the previous positive embeddings $z^+$ (the superscript $(2)$ implies the different distribution compared to $z^+$ in the last subsection).
In particular, for a given $x$, we decide the form of the newly coined positive embedding $z^{(2), +}$ as:
\[
z^{(2), +}=
\begin{cases}
z^+ = h \circ f(x^+),& \text{w.p.}\ \frac12, \\
z_\mathrm{pre} = h \circ f_{\mathrm{pre}}(x),& \text{w.p.}\ \frac12. \\
\end{cases}
\]
With the distribution of the extended positive embedding denoted as $p^{(2)}\paren{z^{(2), +}}$ (the positive pairs $x, x^+$ still follow $p^{(1)}(x, x^+)$),
the proposed \model~loss $\mathcal{L}_{\model}$ can be written as:
\begin{equation}
\label{eqn:contrast_loss}
     \mathcal{L}_{\model} = \underset{\substack{p^{(2)}\paren{z,  z^{(2), +}}  \\ \set{p_n\paren{x_i^-}}_{i=1}^{\abs{\m N_x}}}}{\mathbb{E}}
    \left[-\log \frac{\exp \left(z \cdot  z^{(2), +} / \tau\right)}{\underset{i \in \brkt{\abs{\m N_x}}}{\sum} \exp \left(z \cdot z_{i}^{-} / \tau\right)}\right],
\end{equation}
where $p^{(2)}\paren{z, z^{(2), +}}$ is the joint distribution of $z, z^{(2), +}$ constructed in this subsection. Our proposed $\mathcal{L}_{\model}$ manages to mine the supervised signal at the {\textbf{inter-sample}} level, where we align the positive pairs (composed of different samples) while pushing apart the samples in a negative pool. 









\textcolor{black}{Echoing the findings in \citep{yao2022pcl}, which point out aligning positive pairs across vastly different domains often results in poor performance, our research similarly identifies a substantial gap in the representations of pre-trained and fine-tuned models. Direct alignment using CL as evidenced by our empirical evaluation, tends to be sub-optimal. In response, we introduce the concept of variational generative transformation loss to comprehend the transformation process and bridge these representational gaps. Additionally, the generative transformation module is designed to reconstruct the features of the pre-trained model at an \textbf{intra-sample level}. This complements the inter-sample level supervision provided by contrastive loss. The module, with its associated loss function, intends to provide a more enriched supervised signal, encapsulating crucial within-sample information. In turn, it serves as a pivotal proxy objective that facilitates model anchoring in Eq. \ref{eqn:contrast_loss}.}




To simplify the notation of the transformation, we abuse the previous notation $\{z, z_\mathrm{pre}\}$ for the output embedding from a certain learned/pre-trained model layer,
omitting the corresponding layer denotation. $z_\mathrm{pre}$ is the fixed supervised signal provided by the pre-trained model. 

With the notation $\{z, z_\mathrm{pre}\}$, we introduce the following variational generative model to parameterize the map $g: z \mapsto z_\mathrm{pre}$ relating the representation manifolds formed by (the first several layers of) the learned map $f$ and the fixed pre-trained model $f_\mathrm{pre}$.
In particular, $g$ is composed of an encoder $\phi$ modeling a tunable conditional distribution $q_\phi\paren{z_\mathrm{lat} \mid z}$ of $z_\mathrm{lat}$ and a tunable decoder $\psi$ mapping $z_\mathrm{lat}$ back to $z_\mathrm{pre}$, 
in which $z_\mathrm{lat} \in \mb R^{d'}$ is the latent representation of the generator.
Similar to the training of a regular variational autoencoder (VAE)~\citep{kingma2019introduction}, 
the latent variable $z_\mathrm{lat}$ will be sampled from $q_\phi\paren{z_\mathrm{lat} \mid z}$;
we can then project $z_\mathrm{lat}$ to the pre-trained embedding space via decoder $\psi$ for reconstruction. 
Our variational generative transformation loss $\mathcal{L}_{\model}^{\text{Gen}}$ is designed as:
\begin{align}
\label{eq:gen}
\mathcal{L}_{\model}^{\text{Gen}} = -\mathbb{E}&_{q_\phi\left(z_\mathrm{lat} \mid z\right)}\left[\log p_\psi\left(z_\mathrm{pre} \mid z_\mathrm{lat}\right)\right] \notag\\
&+ \operatorname{KL}\left[q_\phi\left(z_\mathrm{lat} \mid z\right) \parallel p\left(z_\mathrm{lat}\right)\right],
\end{align}
where $p\left(z_\mathrm{lat}\right)$ represents the pre-specified prior distribution of $z_\mathrm{lat}$, $p_\psi\left(z_\mathrm{pre} \mid z_\mathrm{lat} \right)$ is decided by the
``reconstruction loss'' $\|z_\mathrm{pre} - \psi\paren{z_\mathrm{lat}}\|^2$, and the KL divergence term corresponds to the variational regularization term to avoid mode collapse. 
The workflow of our proposed generative transformation is shown in Figure~\ref{fig:gen}. 

Finally, to benefit the representation learning through both generative transformation and our improved contrastive leaning, we set our ultimate objective as:
\begin{equation}
\mathcal{L} = \mathcal{L}_{\text{ERM}}+\lambda \mathcal{L}_{\model}+\beta \mathcal{L}_{\model}^{\text{Gen}}
\end{equation}
where $\lambda$ and $\beta$ are coefficients to balance the multi-task loss. 
The ablation studies in Subsection~\ref{sec:exp ablation} verify the effectiveness of each component.

\section{Experiments}
\label{section:experiments}
\noindent In this section, we empirically evaluate the performance of our proposed \model, intending to answer the following research questions:
\begin{itemize}[itemsep=-0.1em, topsep=0.0em, leftmargin=*]
    \item \textbf{RQ1:} Does \model~enable networks to learn transferable representation under distribution shifts?
    \item \textbf{RQ2:} How do the various components and experimental choices within our \model~ influence the performance?
    \item \textbf{RQ3:} How good is the generalizability of our proposed \model~under different circumstances (e.g., varying label ratios, backbones, modalities)?
    \item \textbf{RQ4:} 
    Does \model~truly establish connections between cross-domain representations?
\end{itemize}

\subsection{Experimental Settings}
{We exhaustively evaluate out-of-domain (OOD) accuracy of \model~on various representative DG benchmarks as in \cite{cha2021swad,yao2022pcl,cha2022domain,chen2022compound}}: OfficeHome \citep{venkateswara2017deep}, PACS \citep{li2017deeper}, VLCS \citep{fang2013unbiased}, TerraIncognita \citep{beery2018recognition}, and DomainNet \citep{peng2019moment}. 
The details of the data sets are shown in Appendix \ref{app:exp_setup}.
{For fair comparison, we strictly follow the experimental settings in \cite{gulrajani2020search,cha2021swad,yao2022pcl,chen2022compound} and adopt the widely used leave-one-domain-out evaluation protocol, i.e., one domain is chosen as the held-out testing domain and the rest are regarded as source training domains.} The experiment results are all averaged over three repeated runs. 
Following DomainBed \citep{gulrajani2020search}, we leave 20\% of source domain data for validation and model selection. 
As in previous works \citep{cha2022domain,yao2022pcl}, we use the ResNet-50 model pre-trained on ImageNet by default, and our code is mainly built upon DomainBed \citep{gulrajani2020search} and SWAD \citep{cha2021swad}. All baselines employ identical pre-trained backbones and dataset splits. We apply the same level of data augmentation across all datasets. Similarly, all baseline comparisons are made using the same pre-trained model and data augmentation techniques. Due to space constraints, detailed implementation and experimental setups are shown in Appendix A.1. {The limitations, attribution of existing assets, and the use of personal data are discussed in Appendix \ref{sec:discussions}.}


\begin{table}[t]
    \centering
    \caption{\tx{Experiments on PACS with ResNet-50. The dataset comprises four domains: Art (A), Cartoon (C), Photo (P), and Sketch (S). In the table, Column A indicates the target domain is A, while the remaining domains are for training.}}
    \label{table:main_pacs}
    \scalebox{0.78}{
    \begin{tabular}{@{} cccccc}
\hline \text { Algorithm } & \text { A } & \text { C } & \text { P } & \text { S } & \text { Avg. } \\
\hline \text {L2A-OT \citep{zhou2020learning} } & 83.3 & 78.2 & 96.2 & 73.6  & 82.8 \\
\text { IRM \citep{arjovsky2019invariant} } & 84.8 & 76.4 & 96.7 & 76.1 & 83.5 \\
\text { MetaReg \citep{balaji2018metareg} } & 87.2 & 79.2 & 97.6 & 70.3 & 83.6 \\
\text { DANN \citep{ganin2016domain} } & 86.4 & 77.4 & 97.3 & 73.5 & 83.7 \\
\text { ERM \citep{vapnik1999overview} } & 85.7 & 77.1 & 97.4 & 76.6 & 84.2 \\
\text { GroupDRO \citep{ganin2016domain} } & 83.5 & 79.1 & 96.7 & 78.3 & 84.4 \\
\text { MTL \citep{blanchard2021domain} } & 87.5 & 77.1 & 96.4 & 77.3 & 84.6 \\
\text { I-Mixup \citep{xu2020adversarial} } & 86.1 & 78.9 & 97.6 & 75.8 & 84.6 \\
\text { MMD \citep{li2018domain} } & 86.1 & 79.4 & 96.6 & 76.5 & 84.7 \\
\text { VREx \citep{krueger2021out} } & 86.0 & 79.1 & 96.9 & 77.7 & 84.9 \\
\text { MLDG \citep{li2018learning} } & 85.5 & 80.1 & 97.4 & 76.6 & 84.9 \\
\text { ARM \citep{zhang2020adaptive} } & 86.8 & 76.8 & 97.4 & 79.3 & 85.1 \\
\text { RSC \citep{huang2020self} } & 85.4 & 79.7 & 97.6 & 78.2 & 85.2 \\
\text { Mixstyle \citep{zhou2021domain} } & 86.8 & 79.0 & 96.6 & 78.5 & 85.2 \\
\text { ER \citep{zhao2020domain} } & 87.5 & 79.3 & \textbf{98.3} & 76.3 & 85.3 \\
\text { pAdaIN \citep{nuriel2021permuted} } & 85.8 & 81.1 & 97.2 & 77.4 & 85.4 \\
\text { SelfReg \citep{kim2021selfreg} } & 85.0 & 81.0 & 95.9 & 80.5 & 85.6 \\
\text { EISNet \citep{wang2020learning} } & 86.6 & 81.5 & 97.1 & 78.1 & 85.8 \\
\text { CORAL \citep{sun2016deep} } & 88.3 & 80.0 & 97.5 & 78.8 & 86.2 \\
\text { SagNet \citep{nam2021reducing} } & 87.4 & 80.7 & 97.1 & 80.0 & 86.3 \\
\text { MADG \citep{dayal2023madg}} & 87.8 & 82.2 & 97.7 & 78.3 & 86.5 \\
\text { DSON \citep{seo2020learning} } & 87.0 & 80.6 & 96.0 & 82.9 & 86.6 \\
\text{SAGM \citep{wang2023sharpness}}  & 87.4 & 80.2 & 98.0 & 80.8 & 86.6 \\
\text{RDM \citep{nguyen2024domain}} & 88.4& 81.3& 97.1& 81.8 & 87.2 \\
\text { COMEN \citep{chen2022compound}} & 88.1 & 82.6 & 97.2 & 81.9 & 87.5 \\
\text { SWAD \citep{cha2021swad} } & 89.3 & 83.4 & 97.3 & 82.5 & 88.1 \\
DRM \citep{zhang2023domain} & 89.6 & 83.4 & 98.4 & 82.3 & 88.4 \\
\text { MIRO \citep{cha2022domain}} & 89.8 & 83.6 & 98.2 & 82.1 & 88.4 \\
\text { PCL \citep{yao2022pcl}} & 90.2 & 83.9 & 98.1 & 82.6 & 88.7 \\
\hline \text { Ours } & \textbf{90.5} & \textbf{84.2} & 98.0 & \textbf{83.3} & \textbf{89.1$\pm$ 0.1} \\
\hline
\end{tabular}}
\label{table:sub_pacs_a}
\vspace{-0.3cm}
\end{table}

\subsection{Results (RQ1)}
We provide comprehensive comparisons with a set of strong baselines on the domain generalization benchmarks PACS and OfficeHome, as shown in Tables \ref{table:sub_pacs_a} and \ref{table:sub_pacs_b}, with results for TerraIncognita, VLCS, and DomainNet datasets deferred to Appendix \ref{app:experiments} due to space limitations. The methods in each table are ranked based on their performance on the dataset. The baselines cover a broad and comprehensive range, including improved learning policies \cite{li2018learning, balaji2018metareg}, enhanced augmentation methods \cite{xu2020adversarial, zhou2020learning}, and domain invariant learning \cite{arjovsky2019invariant, dayal2023madg} from both data \cite{yao2022pcl} and model \cite{cha2022domain} perspectives.

We observe our proposed method achieves the best performance across different kinds of baselines:
the metrics are 44.0 (ERM)$\rightarrow$47.0 (Best Baseline)$\rightarrow$47.5 (Ours) on DomainNet, 77.3$\rightarrow$79.6$\rightarrow$80.0 on VLCS, and 47.8$\rightarrow$52.9$\rightarrow$53.7 on TerraIncognita. 
The results of the intermediate columns in the tables represent performance on the testing domain. For example, ``A'' in Table \ref{table:main_pacs} denotes testing on domain Art and training on Photo, Cartoon, and Sketch. The final result is averaged over all domains. The symbol + in the tables is used to denote that the reproduced experimental performance is clearly distinct from the reported one (such as ``PCL$^{+}$'' in Table~\ref{tab:backbone}). 
All the baselines are sorted in ascending order of their performance. 

We have the following findings from the tables. \ri We find that \model~substantially outperforms all the baseline methods concerning OOD accuracy. This indicates the capability of \model~to extract transferable representation for generalization under distribution shift. \rii We notice most baselines make explicit use of domain supervision, while only a few methods such as RSC \citep{huang2020self}, SagNet \citep{nam2021reducing}, COMEN \citep{chen2022compound}, SWAD \citep{cha2021swad}, MIRO \citep{cha2022domain} and our \model~do not. The excellent performance of our \model~may reveal previous works do not well utilize the domain information and there is still much room for improvement. 
\riii We note that PCL~\citep{yao2022pcl} (Proxy Contrastive Learning) has utilized the potential of CL, aligns embeddings of different samples into domain centers, and consistently achieves good performance.
{Meanwhile, MIRO~\citep{cha2022domain} also preserves the pre-trained features by adding the mutual information regularization term and attains satisfactory performance. 
However, because of their deficiency to connect cross-domain representations, our method manages to improve upon the success the previous baselines had.}


\begin{table}[t]
    \centering
        \caption{Experimental comparisons on Office-Home with state-of-the-art methods on
benchmarks with ResNet-50. }
\scalebox{0.78}{
    \begin{tabular}{cccccc}
\hline Algorithm & A & C & P & R & Avg \\

\hline Mixstyle \citep{zhou2021domain} & $51.1$ & $53.2$ & $68.2$ & $69.2$ & $60.4$ \\
IRM \citep{arjovsky2019invariant} & $58.9$ & $52.2$ & $72.1$ & $74.0$ & $64.3$ \\
ARM \citep{zhang2020adaptive} & $58.9$ & $51.0$ & $74.1$ & $75.2$ & $64.8$ \\
RSC \citep{huang2020self} & $60.7$ & $51.4$ & $74.8$ & $75.1$ & $65.5$ \\
\text {L2A-OT \citep{zhou2020learning} } & 60.6 & 50.1 & 74.8 & 77.0 & 65.6 \\
CDANN \citep{li2018domain} & $61.5$ & $50.4$ & $74.4$ & $76.6$ & $65.7$ \\
DANN \citep{ganin2016domain} & $59.9$ & $53.0$ & $73.6$ & $76.9$ & $65.9$ \\
GroupDRO \citep{ganin2016domain} & $60.4$ & $52.7$ & $75.0$ & $76.0$ & $66.0$ \\
MMD \citep{li2018domain} & $60.4$ & $53.3$ & $74.3$ & $77.4$ & $66.4$ \\
MTL \citep{blanchard2021domain} & $61.5$ & $52.4$ & $74.9$ & $76.8$ & $66.4$ \\
VREx \citep{krueger2021out} & $60.7$ & $53.0$ & $75.3$ & $76.6$ & $66.4$ \\
MLDG \citep{li2018learning} & $61.5$ & $53.2$ & $75.0$ & $77.5$ & $66.8$ \\
\text{RDM \citep{nguyen2024domain}} & 61.1 & 55.1 & 75.7 & 77.3 & 67.3 \\
ERM \citep{vapnik1999overview} & $63.1$ & $51.9$ & $77.2$ & $78.1$ & $67.6$ \\
\text { SelfReg \citep{kim2021selfreg} } & 63.6 & 53.1 & 76.9 & 78.1 & 67.9 \\
I-Mixup \citep{xu2020adversarial} & $62.4$ & $54.8$ & $76.9$ & $78.3$ & $68.1$ \\
SagNet \citep{nam2021reducing} & $63.4$ & $54.8$ & $75.8$ & $78.3$ & $68.1$ \\
CORAL \citep{sun2016deep} & $65.3$ & $54.4$ & $76.5$ & $78.4$ & $68.7$ \\
\text { COMEN \citep{chen2022compound}} & 65.4 & 55.6 & 75.8 & 78.9 & 68.9 \\
\text{SAGM \citep{wang2023sharpness}} & 65.4 & 57.0 & 78.0 & 80.0 & 70.1 \\
SWAD \citep{cha2021swad} & $66.1$ & $57.7$ & $78.4$ & $80.2$ & $70.6$ \\
\text { MADG \citep{dayal2023madg}} & 68.6 & 55.5 & 79.6 & 81.5 & 71.3 \\
\text { PCL \citep{yao2022pcl}} & 67.3 & \textbf{59.9} & 78.7 & 80.7 & 71.6 \\
\text { MIRO \citep{cha2022domain}} & 68.8 & 58.1 & 79.9 & 82.6 & 72.4 \\
\hline Ours & $\mathbf{70.1}$ & $59.1$ & $\mathbf{81.4}$ & $\mathbf{83.4}$ & $\mathbf{73.5\pm 0.2}$ \\
\hline
\end{tabular}}
    \label{table:sub_pacs_b}
\vspace{-0.4cm}
\end{table}

\subsection{Ablation Studies (RQ2)}
\label{sec:exp ablation}
In this part, we investigate the effectiveness of the proposed \model~in Table \ref{tab:ablation} by evaluating the impact of different components. We denote the Cross-Domain Contrastive learning in Section \ref{subsection:strategy} as CDC (with more aggressive data augmentation and cross-domain positive samples), Pre-trained Model Anchoring in Section \ref{subsection:pre-train} as PMA, and Generative Transformation in Eq. \ref{eq:gen} as GT. The ablation results are summarized in Table \ref{tab:ablation}. 
The check mark in the table indicates the module is incorporated. 
{We note that our improved contrastive learning loss in Eqn.~(\ref{eqn:contrast_loss}) has two components: CDC and PMA. 
The overall improvement of the loss is substantial: $70.6 \rightarrow 72.9$.} From the table, we can observe that all the components are useful: when any one of these components is removed, the performance drops accordingly. 
For example, removing PMA module leads to significant performance degeneration, which verifies the importance of anchoring learned maps to pre-trained models. We can then find the combination of PMA and GT leads to the highest improvement in the ablation, which indicates GT and PMA modules complement each other in an effective way. The finding is also consistent with our motivation for generative transformation loss. 
Moreover, we also evaluate self-contrastive learning. The experimental results indicate that self-contrastive learning will distort the learned embeddings and hamper performance. Besides, the experiment without aggressive data augmentation also validates the effectiveness of stronger data augmentations we suggest in Section \ref{subsection:strategy}. In this paper, we increase the intensity of data augmentation operations beyond what is used in typical supervised learning to achieve more aggressive data augmentation. More details and further experimental verification can be found in Table \ref{tab:aug_ablate} in the Appendix. The efficiency and impact of hyper-parameters are shown in Appendix \ref{app:efficiency} and \ref{app:hyper}. We note that our method exhibits similar or even lower time and memory costs while stably outperforming baselines regardless of different hyper-parameters. {Additional experimental details and explanations regarding our choices for VAE structures, contrastive learning techniques within \model, cross-domain examples in CDC, and the Wilds Benchmark can be found in Appendix \ref{app:ablation}. The experimental results further verify the robustness of our proposed \model.}





\begin{table}[t]
\centering
\caption{Ablation Studies of \model~on OfficeHome.}
\vspace{-0.2cm}
\label{tab:ablation}
\scalebox{0.93}{
\begin{tabular}{ccc|cccc|c}
\hline
CDC  &  PMA  & GT    & A                & C             & P             & R             & Avg.          \\
\hline
-             & -             & -             & 66.1          & 57.7          & 78.4          & 80.2          & 70.6          \\
\multicolumn{3}{c|}{with Self-Contrast}                 & 65.4          & 51.4          & 79.1          & 79.5          & 68.9          \\
\hline
$ \checkmark$ & -             & -             & 68.0          & 57.9          & 80.1          & 81.3          & 71.8          \\
-             & $ \checkmark$ & -             & 68.8          & 57.8          & 80.4          & 82.3          & 72.3          \\
-             & -             & $ \checkmark$ & 69.0          & 56.9          & 80.6          & 81.6          & 72.0          \\
-             & $ \checkmark$ & $ \checkmark$ & 70.0          & 58.7          & 80.5          & 83.4          & 73.1          \\
$ \checkmark$ & $ \checkmark$ & -             & 69.2          & 58.5          & 81.0          & 83.0          & 72.9          \\
$ \checkmark$ & -             & $ \checkmark$ & 69.0          & 58.5          & 80.7          & 82.1          & 72.6          \\
\hline
\multicolumn{3}{c|}{w/o Aggressive Aug}        & 69.8          & 58.6          & 81.0          & 82.6          & 73.0          \\
$ \checkmark$ & $ \checkmark$ & $ \checkmark$ & \textbf{70.1} & \textbf{59.1} & \textbf{81.4} & \textbf{83.4} & \textbf{73.5} \\
\hline
\end{tabular}
}
\vspace{-0.2cm}
\end{table}

\begin{table}[t]

\centering
\caption{Experimental comparisons of \model~with representative baselines on OfficeHome under various label ratios.}
\vspace{-0.2cm}
\label{tab:ratio}
\scalebox{0.85}{
\begin{tabular}{ccccccc}
\hline
          Ratio           & Algorithm                              & A             & C             & P             & R             & Avg.                  \\ \hline
\multirow{6}{*}{5\%}  & ERM \citep{vapnik1999overview}          & 40.4          & 32.6          & 42.6          & 49.2          & 41.2                  \\
                     & SWAD \citep{cha2021swad}                & 46.9          & 36.2          & 48.5          & 54.2          & 46.4                  \\
                     & \text{COMEN} \citep{chen2022compound}   & 47.7          & 39.2          & 50.6          & 56.1          & 48.4                  \\
                     & \text{ $\text{PCL}$ }\citep{yao2022pcl} & 48.4          & 42.3          & 55.2          & 57.2          & 50.8                  \\
                     & \text { MIRO }\citep{cha2022domain}     & 51.0          & 41.6          & 58.6          & 61.5          & 53.2                  \\ \cline{2-7} 
                     & Ours                                   & \textbf{55.7} & \textbf{44.1} & \textbf{63.1} & \textbf{67.1} & \textbf{57.5 (+16.3)} \\ \hline
\multirow{6}{*}{10\%} & ERM \citep{vapnik1999overview}          & 45.1          & 41.9          & 55.9          & 58.0          & 50.2                  \\
                     & \text{COMEN} \citep{chen2022compound}   & 50.4          & 44.3          & 56.8          & 60.9          & 53.1                  \\
                     & SWAD \citep{cha2021swad}                & 53.3          & 43.9          & 61.8          & 65.2          & 56.1                  \\
                     & \text{ $\text{PCL}$ }\citep{yao2022pcl} & 54.6          & 45.1          & 60.9          & 67.2          & 57.0                  \\
                     & \text { MIRO }\citep{cha2022domain}     & 58.9          & 46.6          & 68.6          & 71.7          & 61.4                  \\ \cline{2-7} 
                     & Ours                                   & \textbf{62.5} & \textbf{49.2} & \textbf{72.3} & \textbf{75.1} & \textbf{64.8 (+14.6)} \\ \hline
\end{tabular}}
\vspace{-0.3cm}
\end{table}

\subsection{Case Studies}
\label{sec:exp_case}
\textbf{Generalization ability (RQ3)}.
To verify the generalizability of our proposed \model, we first conduct experiments\footnote{We select a few of the most representative methods as baselines.} 
with different label ratios (the percentage of labeled training data) and backbones. 
\ri In Table \ref{tab:ratio}, we find \model~can obtain consistent improvement over baselines, in both cases of 5\% and 10\% label ratios. Our method yields a 16.3 and 14.6 absolute improvement compared with ERM. We can observe that as the number of available labels reduces, the model benefits more from our \model~(compared with previous 67.6$\rightarrow$73.5 increase under 100\% label ratio in Table \ref{table:sub_pacs_b}).
\rii In Table \ref{tab:backbone}, we test the performance with a new backbone, ResNet-18 (previously ResNet-50)\footnote{For semantic information matching, pre-trained representations in \model~are generated from the same backbone model used for fine-tuning.}. 
We find that even though the baselines' relative ordering changes significantly, our model still performs the best, showcasing the robustness thereof. We further observe replacing the ResNet-18 pre-trained representations to the larger ResNet-50 ones (``mismatch'' between the backbone used for fine-tuning and the pre-trained representations) will cause substantial performance drop $67.5 \rightarrow 62.6$. The superior performance of \model~on more backbones (RegNet, ViT) are shown in Table \ref{tab:source} in Appendix.


\textbf{Analysis of the representations in \model~(RQ4)}.
Here we analyze the representations in \model~to provide more insights. 
In Figure \ref{fig:cross_domain}, we utilize t-SNE \citep{van2008visualizing} to visualize the embeddings in the pre-trained model, ERM, SCL and our \model. 
We observe that mapped by the original pre-trained model ResNet-50, the intra-class samples of the training domains and the testing domains are scattered while well-connected. 
However, in the ERM model, many samples in the testing domain are distributed in the central part of the plot, which is separated from the training samples. There is a clear gap between the training and the testing domains. As for SCL, it seems to harm the learned embedding space and distort the class decision boundary. Our proposed \model~can effectively cluster the intra-class samples across domains. We then visualize the embeddings in ERM, PCL, and our \model~ on the testing domains in \Cref{app:visualization}. Our \model~learns discriminative representations even in the unseen target domain by enhancing \textcolor{black}{intra-class connectivity}, which is unaddressed in ERM and PCL.


\begin{table}[t]
\centering
\caption{Experimental comparisons of \model~on OfficeHome with the ResNet-18 backbone in use.}
\vspace{-0.2cm}
\label{tab:backbone}
\begin{tabular}{cccccc}
\hline
Algorithm                                                                               & A               & C                              & P               & R               & Avg.            \\ \hline
ERM \citep{vapnik1999overview}                                      & 50.6            & 49.0                           & 69.9            & 71.4            & 60.2            \\
SWAD \citep{cha2021swad}                                           & 54.6            & 50.0                           & 71.1            & 72.8            & 62.1            \\
\text{ $\text{PCL}$ }\citep{yao2022pcl} & 58.8            & 51.9                           & 74.2            & 75.2            & 65.0            \\
\text { MIRO }\citep{cha2022domain}                    & 59.7            & 52.6                           & 75.0            & 77.7            & 66.2            \\
\text { COMEN }  \citep{chen2022compound}              & 57.6            & \textbf{55.8} & 75.5            & 76.9            & 66.5            \\ \hline

``Mismatch''         & 53.4 & 50.7                        & 72.3 & 74.0 & 62.6        \\
Ours                                                                                    & \textbf{61.7} & 53.6                         & \textbf{75.9} & \textbf{78.7} & \textbf{67.5}

\\ \hline
\end{tabular}
\vspace{-0.2cm}
\end{table}



\section{Related work}
\label{section:related work}

In this section, we review the related works in domain generalization and contrastive learning.

\subsection{Domain Generalization}
Improving model robustness under distribution shifts has also been extensively studied in domains such as recommender systems~\citep{causal,wei2024towards,wei2021model,wei2020fast,wei2022comprehensive,ban2021ee}, federated learning~\citep{bao2024boba,bao2024adaptive}, and graph learning~\citep{chen2023coarsening,bao2024adarc,zhou2024graph,yan2024reconciling,lin2024bemap,linbacktime,wang2024made}.
The goal of DG is to enable models to generalize to unknown target domains under distribution shifts. The related literature can be split into several categories as follows. 

(i) The first line of work focuses on learning policies.
One strategy is meta learning \citep{finn2017model}, which adapts to new environments rapidly with limited observations; 
the meta-optimization idea was thus introduced in DG \citep{li2018learning,balaji2018metareg,qiao2020learning} to generalize to future testing environments/domains; 
another widely-studied strategy is ensemble learning~\citep{cha2021swad,chu2022dna}, claiming DG can benefit from several diverse neural networks to obtain more robust representations. 
(ii) The second line of work is data augmentation. Many fabricated or learnable augmentation strategies \citep{volpi2018generalizing,zhou2020learning,li2021simple,xu2020adversarial} were developed to regularize and enhance deep learning models. In our paper, we verify more aggressive augmentation can lead to better representations in CL as well. 
(iii) The last series of work is domain invariant learning. Researchers seek to learn invariances across multiple observed domains for improved generalization on target domains. The commonly used approaches include domain discrepancy regularization \citep{li2018domain,zhou2020domain} and domain adversarial learning \citep{li2018deep,ganin2016domain,matsuura2020domain}. 
Recently, MIRO~\citep{cha2022domain} began to explore the retention of pre-trained features by designing the mutual information regularization term. \textcolor{black}{The paper \citep{liu2023promoting} also utilized the concept connectivity to build up the method. However, their concept of "connectivity" based on joint distribution clearly differ from our paper. Therefore the theoretical motivation behind two papers are indeed different. Moreover, the methods proposed are different. Except for the common strategy of strong augmentation recommended by the contrastive learning theory paper \cite{wang2022chaos}, our proposed methods are different from the ones in \cite{liu2023promoting}. They propose two nearest-neighbor-based methods for constructing positive pairs, while our main contribution lies in the exploitation of both the pre-trained models and the intra-class data connectivity.}


\subsection{Contrastive Learning}

Contrastive learning (CL) \citep{chen2020simple} aims to learn discriminative sample representation by aligning positive instances and pushing negative ones apart. As a promising self-supervised learning paradigm, CL is widely used in unsupervised pre-training to improve the performance of downstream tasks~\citep{hjelm2018learning,gao2021simcse,wei2022augmentations,li2022let,he2020momentum,chen2020simple,caron2020unsupervised,chen2021exploring,grill2020bootstrap, he2023robust}. 
SimCLR~\citep{chen2020simple} is the CL framework that first reveals the projection head and data augmentation as the core components to learn invariant representation across views. MoCo~\citep{he2020momentum} proposes to build a dynamic queue dictionary to enlarge batch size for effective learning. There are also works~\citep{khosla2020supervised,gunel2020supervised,cui2021parametric} adapting CL to the supervised setting to leverage label information.

The capability of CL to obtain class-separated representations has also motivated the application in domain generalization.
SelfReg~\citep{kim2021selfreg} introduced a new regularization method to build self-supervised signals with only positive samples; 
PCL~\citep{yao2022pcl} proposed a proxy-based approach to alleviate the positive alignment issue in CL; COMEN~\citep{chen2022compound} used a prototype-based CL component to learn the relationships between various hidden clusters.
However, the role of CL in domain generalization is not yet well explored, and our work is dedicated to shedding some light on the understanding of its effect from a \textcolor{black}{intra-class connectivity} perspective.




\section{Conclusions}
\label{section:conclusion}
In this paper, we revisit the role of contrastive learning (CL) in domain generalization and identify a key factor: \textcolor{black}{intra-class connectivity}. We further realize this characteristic of representations can be attained from two aspects, data and model.
On the data side, we analyze the failure of directly applying CL to DG and propose two strategies to improve \textcolor{black}{intra-class connectivity}: \ri applying more aggressive data augmentation and \rii expanding the scope of positive samples.
On the model side, to alleviate lack of access to the testing domains in training, we propose to anchor learned maps to pre-trained models which enhances the desired connectivity between training and testing domains. Generative transformation is further introduced to complement the pre-trained alignment. 
Consequently, we combine the pieces together and propose \model~to enable robust representations in the out-of-domain scenario. 
Extensive experiments on five real-world datasets demonstrate the effectiveness of \model, which outperforms a bundle of baselines.

\begin{acks}
This work is supported by National Science Foundation under Award No. IIS-2117902, and Agriculture and Food Research Initiative (AFRI) grant no. 2020-67021-32799/project accession no.1024178 from the USDA National Institute of Food and Agriculture. The views and conclusions are those of the authors and should not be interpreted as representing the official policies of the funding agencies or the government.
\end{acks}



\bibliography{dccl}
\balance
\bibliographystyle{ACM-Reference-Format}

\clearpage
\appendix
\onecolumn


\section{Details of experiments}
\label{sec:details_exp}

\subsection{Experimental Setup}
\label{app:exp_setup}

\begin{table}[htbp]
\centering
\caption{Statistics of datasets.}
\label{tab:stat}
\begin{tabular}{cccc}
\hline
     Datasets          & \# images & \# domains & \# classes \\
\hline
PACS           & 9991     & 4         & 7         \\
VLCS           & 10729    & 4         & 5         \\
OfficeHome     & 15588    & 4         & 65        \\
TerraIncognita & 24788    & 4         & 10        \\
DomainNet      & 586575   & 6         & 345      \\
\hline
\end{tabular}
\end{table}

Here we elaborate the detailed experimental setup of our paper. 
{Following DomainBed \citep{gulrajani2020search}, we split 80\%/20\% data from source domains as the training/validation set. The best-performing model on the validation set will be evaluated on the testing target domain to obtain the test performance.} 
The statistics of the experimental datasets are shown in Table \ref{tab:stat}. We list the number of images, domains, classes in each dataset. The proposed model is optimized using Adam \citep{kingma2014adam} with the learning rate of 5e-5. The hyper-parameter $\lambda$ is searched over \{0.1, 1, 2, 5\}, and $\beta$ is tuned in the range of \{0.01, 0.05, 0.1\}. The temperature $\tau$ is set to 0.1 by default. For the projection head used for contrastive learning, we use a two-layer MLP with ReLU and BatchNorm. Regarding variational reconstruction, following \cite{cha2022domain}, we employ a simple yet effective architecture, in which the identity function is used as mean encoder and a bias-only network with softplus activation for the variance encoder. More intricate architecture can be explored in the future. Following \cite{gulrajani2020search}, for all the datasets except DomainNet, we train the model for 5000 steps. For the DomainNet dataset, we train the model for 15000 steps. Other algorithm-agnostic hyper-parameters such as the batch size are all set to be the same as in the standard benchmark DomainBed \citep{gulrajani2020search}. 
{For batch construction, we sample the same number of samples from each training domain as in DomainBed \citep{gulrajani2020search}. Generative Transformation is done for all 4 layers in ResNet-18/50.} 
The experiments are all conducted on one Tesla V100 32 GB GPU. The baseline results are taken from the original papers. If the results were not available, we reproduced them for fair comparisons. For the data augmentation strategy, previous works usually adopted random cropping, grayscale, horizontal flipping and random color jittering. In this paper, we simply increase the intensity of random color jittering to achieve more aggressive data augmentation on all datasets. 
Developing stronger and more adaptive augmentation methods for contrastive learning on DG may further enhance the performance.

\subsection{Experimental Results on TerraIncognita, VLCS, and DomainNet Data Sets}
\label{app:experiments}
We put the experimental comparisons with state-of-the-art baselines on TerraIncognita, VLCS, and DomainNet data sets respectively in Tables~\ref{table:main_terra}, \ref{table:main_vlcs}, and \ref{table:main_domain}. The symbol + in the tables is used to denote that the reproduced experimental performance is distinct from the originally reported one such as ``PCL$^{+}$'' in Table \ref{table:main_domain}. 
We can observe our proposed \model~still surpasses previous methods, which is consistent with the conclusion in the main text and successfully verify the effectiveness of our proposed method.

\subsection{Visualization}
\label{app:visualization}
\begin{figure*}[ht]
     \centering
     \begin{subfigure}[b]{0.32\textwidth}
         \centering
         \includegraphics[width=\textwidth]{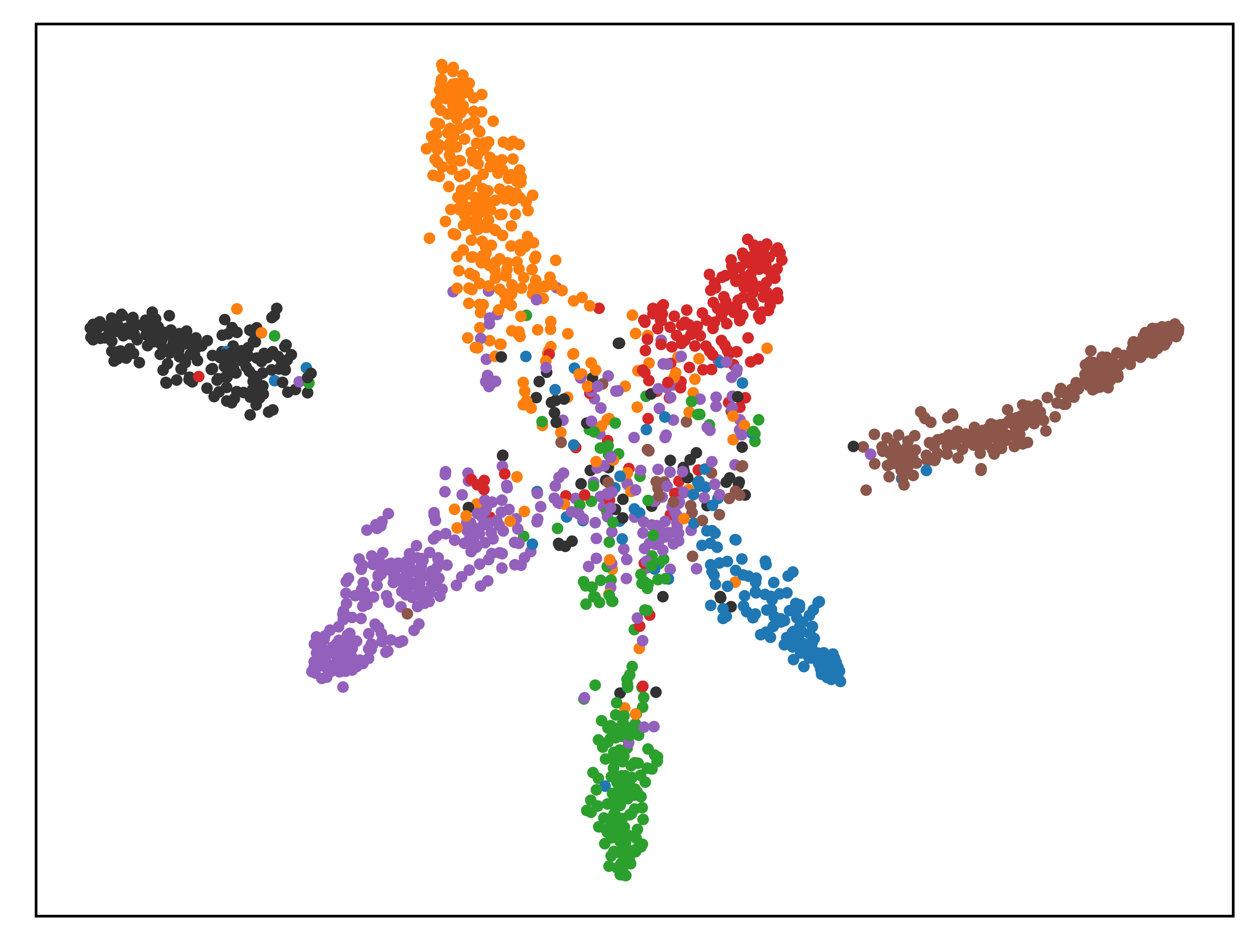}
         \caption{ERM}
         \label{fig:erm}
     \end{subfigure}
     \hfill
     \begin{subfigure}[b]{0.32\textwidth}
         \centering
         \includegraphics[width=\textwidth]{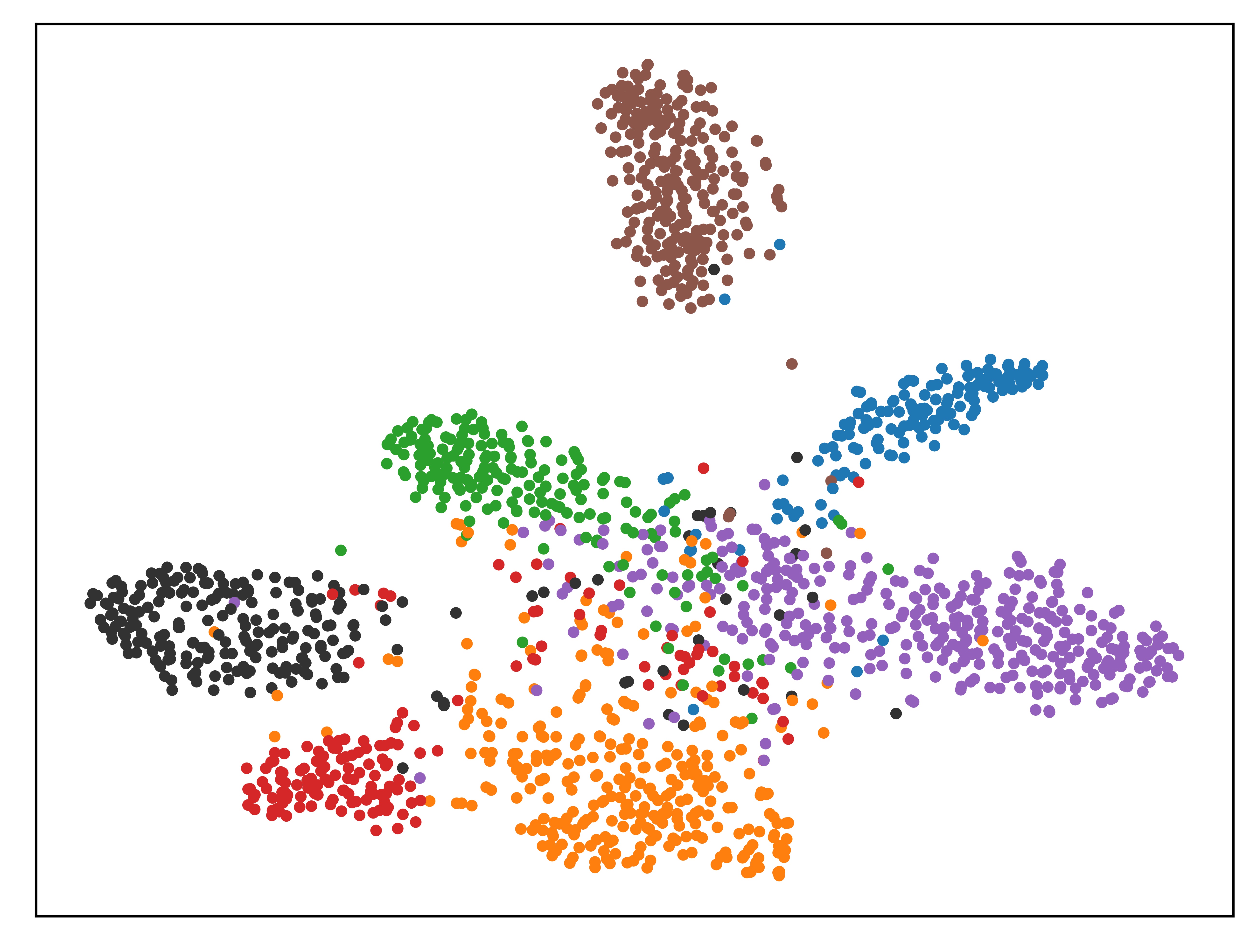}
         \caption{PCL}
         \label{fig:pcl}
     \end{subfigure}
     \begin{subfigure}[b]{0.32\textwidth}
         \centering
         \includegraphics[width=\textwidth]{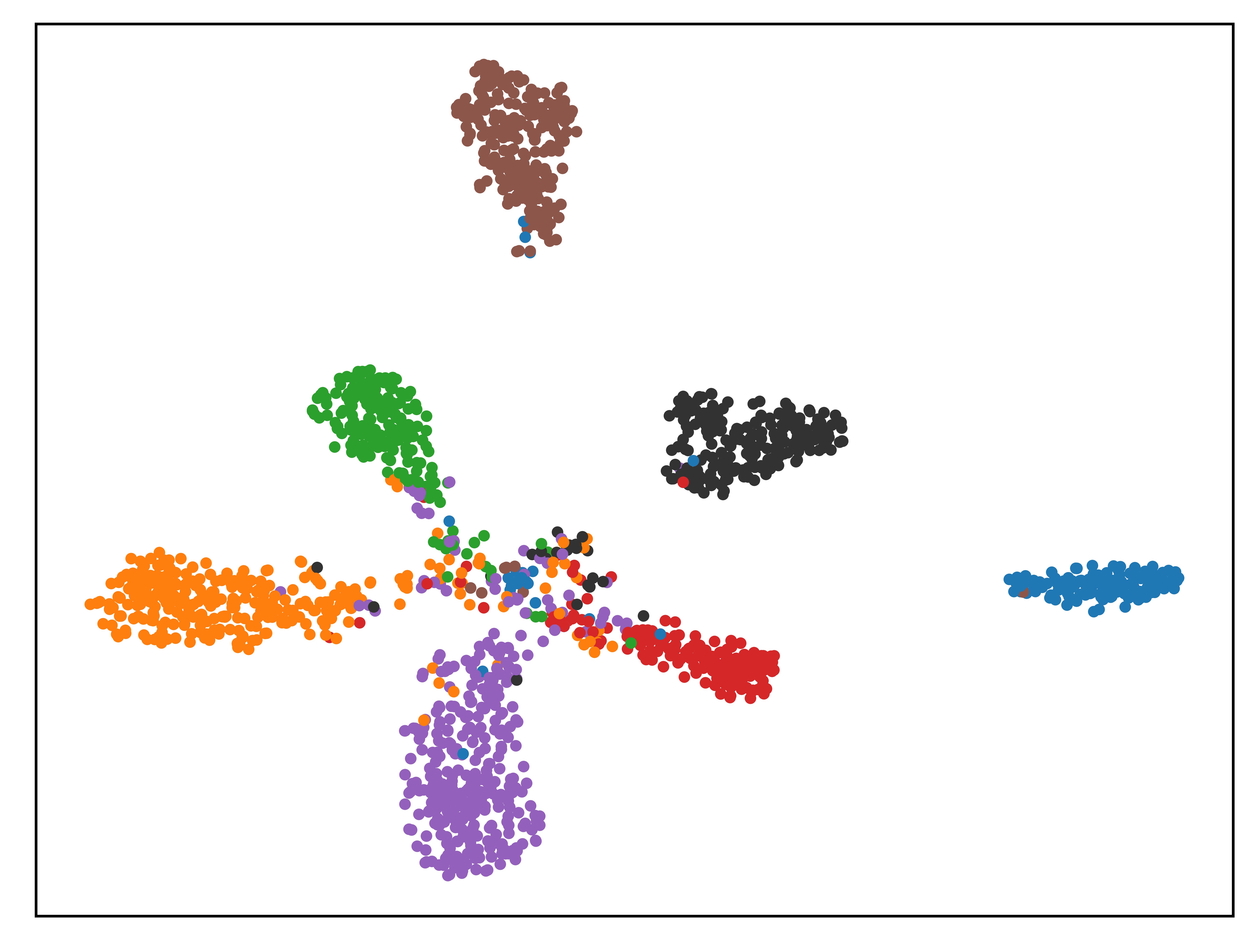}
         \caption{\model}
         \label{fig:dccl}
     \end{subfigure}
        \caption{t-SNE visualization of the ERM, PCL and \model~representations on the testing domain. Same-class points are in the same colors.
        We visualize the embedding on PACS dataset where the source domains are photo, sketch, and cartoon; the target domain is art.}
        \label{fig:model}
\end{figure*}

We demonstrate the embeddings of ERM, PCL, and our \model~methods on the testing domain in Figure \ref{fig:model}. 
ERM, among the three methods, has the most samples distributed in the central area which cannot be distinguished. For the embedding of contrastive-learning-based baseline PCL, there are fewer samples distributed ambiguously. However, the class clusters are not compact and the class boundaries are not clear. 
By contrast, our \model~learns discriminative representations even in the unseen target domain by enhancing \textcolor{black}{intra-class connectivity} in CL.


\begin{table}[t]
\centering
\caption{Experimental comparisons with state-of-the-art methods on TerraIncognita
benchmark with ResNet-50.}
\vspace{-0.2cm}
\label{table:main_terra}
\scalebox{0.9}{
\begin{tabular}{cccccc}
\hline Algorithm & L100 & L38 & L43 & L46 & Avg. \\
\hline MMD \citep{li2018domain} & $41.9$ & $34.8$ & $57.0$ & $35.2$ & $42.2$ \\
GroupDRO \citep{ganin2016domain} & $41.2$ & $38.6$ & $56.7$ & $36.4$ & $43.2$ \\
Mixstyle \citep{zhou2021domain} & $54.3$ & $34.1$ & $55.9$ & $31.7$ & $44.0$ \\
ARM \citep{zhang2020adaptive} & $49.3$ & $38.3$ & $55.8$ & $38.7$ & $45.5$ \\
MTL \citep{blanchard2021domain} & $49.3$ & $39.6$ & $55.6$ & $37.8$ & $45.6$ \\
CDANN \citep{li2018domain} & $47.0$ & $41.3$ & $54.9$ & $39.8$ & $45.8$ \\
VREx \citep{krueger2021out} & $48.2$ & $41.7$ & $56.8$ & $38.7$ & $46.4$ \\
RSC \citep{huang2020self} & $50.2$ & $39.2$ & $56.3$ & $40.8$ & $46.6$ \\
DANN \citep{ganin2016domain} & $51.1$ & $40.6$ & $57.4$ & $37.7$ & $46.7$ \\
SelfReg \citep{kim2021selfreg} & $48.8$ & $41.3$ & $57.3$ & $40.6$ & $47.0$ \\
RDM \citep{nguyen2024domain} & 52.9 & 43.1 & 58.1 & 36.1 & 47.5 \\
IRM \citep{arjovsky2019invariant} & $54.6$ & $39.8$ & $56.2$ & $39.6$ & $47.6$ \\
CORAL \citep{sun2016deep} & $51.6$ & $42.2$ & $57.0$ & $39.8$ & $47.7$ \\
MLDG \citep{li2018learning} & $54.2$ & $44.3$ & $55.6$ & $36.9$ & $47.8$ \\
ERM \citep{vapnik1999overview} & $54.3$ & $42.5$ & $55.6$ & $38.8$ & $47.8$ \\
I-Mixup \citep{xu2020adversarial} & $59.6$ & $42.2$ & $55.9$ & $33.9$ & $47.9$ \\
SagNet \citep{nam2021reducing} & $53.0$ & $43.0$ & $57.9$ & $40.4$ & $48.6$ \\
\text{SAGM \citep{wang2023sharpness}} & 54.8 & 41.4 & 57.7 & 41.3 & 48.8 \\
\text { COMEN \citep{chen2022compound}} & 56.0 & 44.3 & 58.4 & 39.4 & 49.5 \\
SWAD \citep{cha2021swad} & $55.4$ & $44.9$ & $59.7$ & $39.9$ & $50.0$ \\
\text { PCL \citep{yao2022pcl}} & $58.7$ & $46.3$ & $60.0$ & $43.6$ & $52.1$ \\
\text{MADG \citep{dayal2023madg}} & 59.8 & 50.3 & 57.2 & 42.5 & 52.7 \\
\text { MIRO \citep{cha2022domain}} & 60.9 & 47.6 & 59.5 & 43.4 & 52.9 \\
\hline Ours & \textbf{62.2} & $\mathbf{48.3}$ & $\mathbf{60.6}$ & $\mathbf{4 3.6}$ & $\mathbf{53.7\pm 0.2}$ \\
\hline
\end{tabular}}
\vspace{-0.2cm}
\end{table}

\begin{table}[ht]
\centering
\caption{Experimental comparisons with state-of-the-art methods on VLCS
benchmark with ResNet-50.}
\label{table:main_vlcs}
\begin{tabular}{cccccc}
\hline Algorithm & C & L & S & V & Avg \\
\hline GroupDRO \citep{ganin2016domain} & $97.3$ & $63.4$ & $69.5$ & $76.7$ & $76.7$ \\
RSC \citep{huang2020self} & $97.9$ & $62.5$ & $72.3$ & $75.6$ & $77.1$ \\
MLDG \citep{li2018learning} & $97.4$ & $65.2$ & $71.0$ & $75.3$ & $77.2$ \\
MTL \citep{blanchard2021domain} & $97.8$ & $64.3$ & $71.5$ & $75.3$ & $77.2$ \\
ERM \citep{vapnik1999overview} & $98.0$ & $64.7$ & $71.4$ & $75.2$ & $77.3$ \\
I-Mixup \citep{xu2020adversarial} & $98.3$ & $64.8$ & $72.1$ & $74.3$ & $77.4$ \\
MMD \citep{li2018domain} & $97.7$ & $64.0$ & $72.8$ & $75.3$ & $77.5$ \\
CDANN \citep{li2018domain} & $97.1$ & $65.1$ & $70.7$ & $77.1$ & $77.5$ \\
ARM \citep{zhang2020adaptive} & $98.7$ & $63.6$ & $71.3$ & $76.7$ & $77.6$ \\
SagNet \citep{nam2021reducing} & $97.9$ & $64.5$ & $71.4$ & $77.5$ & $77.8$ \\
SelfReg \citep{kim2021selfreg} & $96.7$ & $65.2$ & $73.1$ & $76.2$ & $77.8$ \\
Mixstyle \citep{zhou2021domain} & $98.6$ & $64.5$ & $72.6$ & $75.7$ & $77.9$ \\
\text { PCL \citep{yao2022pcl}} & 99.0 & 63.6 & 73.8 & 75.6 & 78.0 \\
VREx \citep{krueger2021out} & $98.4$ & $64.4$ & $74.1$ & $76.2$ & $78.3$ \\
RDM \citep{nguyen2024domain} & 98.1 & 64.9 & 72.6 & 77.9 & 78.4 \\
\text { COMEN \citep{chen2022compound}} & 98.5 & 64.1 & 74.1 & 77.0 & 78.4 \\
IRM \citep{arjovsky2019invariant} & $98.6$ & $64.9$ & $73.4$ & $77.3$ & $78.6$ \\
DANN \citep{ganin2016domain} & $99.0$ & $65.1$ & $73.1$ & $77.2$ & $78.6$ \\ 
\text{MADG \citep{dayal2023madg}} & 98.5 & 65.8 & 73.1 & 77.3 & 78.7 \\
CORAL \citep{sun2016deep} & $98.3$ & \textbf{66.1} & $73.4$ & $77.5$ & $78.8$ \\
SWAD \citep{cha2021swad} & $98.8$ & $63.3$ & $75.3$ & $79.2$ & $79.1$ \\
\text{DRM \citep{zhang2023domain}} & 98.8 & 64.3 & 75.0 & 79.9 & 79.5 \\
\text{SAGM \citep{wang2023sharpness}} & 98.6 & 64.1 & 75.1 & 80.2 & 79.5 \\
\text { MIRO \citep{cha2022domain}} & 98.8 & 64.2 & 75.5 & 79.9 & 79.6 \\
\hline Ours & $\mathbf{99.1}$ & $64.0$ & $\mathbf{76.1}$ & $\mathbf{80.7}$ & $\mathbf{80.0\pm 0.1}$ \\
\hline
\end{tabular}
\end{table}

\begin{table}[ht]
\centering
\caption{Experimental comparisons with state-of-the-art methods on DomainNet
benchmark with ResNet-50.}
\label{table:main_domain}
\scalebox{0.85}{
\begin{tabular}{cccccccc}
\hline Algorithm & clip & info & paint & quick & real & sketch & Avg \\
\hline MMD \citep{li2018domain} & $32.1$ & $11.0$ & $26.8$ & $8.7$ & $32.7$ & $28.9$ & $23.4$ \\
GroupDRO \citep{ganin2016domain} & $47.2$ & $17.5$ & $33.8$ & $9.3$ & $51.6$ & $40.1$ & $33.3$ \\
VREx \citep{krueger2021out} & $47.3$ & $16.0$ & $35.8$ & $10.9$ & $49.6$ & $42.0$ & $33.6$ \\
IRM \citep{arjovsky2019invariant} & $48.5$ & $15.0$ & $38.3$ & $10.9$ & $48.2$ & $42.3$ & $33.9$ \\
Mixstyle \citep{zhou2021domain} & $51.9$ & $13.3$ & $37.0$ & $12.3$ & $46.1$ & $43.4$ & $34.0$ \\
ARM \citep{zhang2020adaptive} & $49.7$ & $16.3$ & $40.9$ & $9.4$ & $53.4$ & $43.5$ & $35.5$ \\
CDANN \citep{li2018domain} & $54.6$ & $17.3$ & $43.7$ & $12.1$ & $56.2$ & $45.9$ & $38.3$ \\
DANN \citep{ganin2016domain} & $53.1$ & $18.3$ & $44.2$ & $11.8$ & $55.5$ & $46.8$ & $38.3$ \\
RSC \citep{huang2020self} & $55.0$ & $18.3$ & $44.4$ & $12.2$ & $55.7$ & $47.8$ & $38.9$ \\
I-Mixup \citep{xu2020adversarial} & $55.7$ & $18.5$ & $44.3$ & $12.5$ & $55.8$ & $48.2$ & $39.2$ \\
MADG \citep{dayal2023madg} & 62.5 & 22.0 & 34.1 & 15.1 & 57.4 & 48.0 & 39.9 \\
SagNet \citep{nam2021reducing} & $57.7$ & $19.0$ & $45.3$ & $12.7$ & $58.1$ & $48.8$ & $40.3$ \\
MTL \citep{blanchard2021domain} & $57.9$ & $18.5$ & $46.0$ & $12.5$ & $59.5$ & $49.2$ & $40.6$ \\
MLDG \citep{li2018learning} & $59.1$ & $19.1$ & $45.8$ & $13.4$ & $59.6$ & $50.2$ & $41.2$ \\
CORAL \citep{sun2016deep} & $59.2$ & $19.7$ & $46.6$ & $13.4$ & $59.8$ & $50.1$ & $41.5$ \\
DRM \citep{zhang2023domain} & $60.3$ & $22.0$ & $49.2$ & $13.0$ & $60.5$ & $51.2$ & $42.7$ \\
SelfReg \citep{kim2021selfreg} & $60.7$ & $21.6$ & $49.4$ & $12.7$ & $60.7$ & $51.7$ & $42.8$ \\
RDM \citep{nguyen2024domain} & 62.1 & 20.7 & 49.2 & 14.1 & 63.0 & 51.4 & 43.4 \\
MetaReg \citep{balaji2018metareg} & $59.8$ & $\mathbf{2 5 . 6}$ & $50.2$ & $11.5$ & $64.6$ & $50.1$ & $43.6$ \\
DMG \citep{chattopadhyay2020learning} & $65.2$ & $22.2$ & $50.0$ & $15.7$ & $59.6$ & $49.0$ & $43.6$ \\
ERM \citep{vapnik1999overview} & $63.0$ & $21.2$ & $50.1$ & $13.9$ & $63.7$ & $52.0$ & $44.0$ \\
\text { COMEN \citep{chen2022compound}} & 64.0 & 21.1 & 50.2 & 14.1 & 63.2 & 51.8 & $44.1$ \\
\text{SAGM \citep{wang2023sharpness}} & 64.9 & 21.1 & 51.5 & 14.8 & 64.1 & 53.6 & 45.0\\
\text { $\text{PCL}^{+}$ \citep{yao2022pcl}} & 64.3 & 20.9 & 52.7 & \textbf{16.7} & 62.2 & $55.5$ & $45.4$  \\
\text{MODE-A \citep{dai2023moderately}} & 68.3 & 23.4 & 52.4 & 16.8 & 63.0 & 54.0 & 46.3 \\
SWAD \citep{cha2021swad} & $66.0$ & $22.4$ & $53.5$ & 16.1 & $65.8$ & $55.5$ & $46.5$ \\

\text { MIRO \citep{cha2022domain}} & 66.4 & 23.5 & 54.1 & 16.2 & 66.8 & $54.8$ & $47.0$ \\
\hline Ours & $\mathbf{66.9}$ & $23.0$ & $\mathbf{5 5 . 1}$ & $16.0$ & $\mathbf{6 7. 7}$ & $\mathbf{5 6 . 1}$ & $\mathbf{47.5\pm 0.0}$ \\
\hline
\end{tabular}
}
\end{table}

\subsection{Representation Connectivity of Pre-Trained Models}
\label{app:connectivity}
{Our motivation to utilize pre-trained models for better connectivity is intuitive: we consider pre-trained model can return effective representations modeling the pairwise interactions among images, which thus draws target domains closer to source domains. 
To verify the motivation, we conduct experiments to evaluate whether the pre-trained model is ``well-connected''.}
\begin{enumerate}[itemsep=-0.3em, topsep=-0.05em, leftmargin=*]
\item We design a quantitative \textbf{metric to help evaluate} whether the pre-trained space is ``well-connected''.
For images within the same class, we take those images as nodes and construct a graph, only connecting two nodes when their distance on the pre-trained space is smaller than a threshold.
We denote the smallest possible threshold which makes the graph \textbf{connected} as $\tau$, and denote the mean and the std of the pairwise distances respectively as $\mu$ and $\sigma$.
We can thus use $(\tau - \mu)/\sigma$ as a metric to describe the connectivity of the representations.
\item We report the mean (max) metrics (the smaller, the better) of each class for ERM and pre-trained model on PACS, VLCS, and Terra.; the values for ERM are 1.37 (2.68), 1.78 (2.15), and 3.31 (3.56), for pre-trained model 0.54 (0.81), 0.46 (0.62), and 0.63 (0.76). The results confirm the pre-trained space is well-connected.
\end{enumerate} 
Furthermore, the \underline{variation in performance improvement} across different datasets can be attributed to differences in connectivity. We define a measure to evaluate connectivity in Appendix A.4 where lower values indicate better connectivity. For the pre-trained (ERM) model, the connectivity measure we have is 0.54 (1.37) for PACS and 0.49 (2.85) for OfficeHome. A larger discrepancy in connectivity between ERM and the pretraine model ($\frac{1.37}{0.54}$ v.s. $\frac{2.85}{0.49}$) allows for greater potential for improvement.

\subsection{Further Ablation Study}
\label{app:ablation}
\textbf{Choices of VAE structures}.
In our experiments, using more advanced VAE structures like HFVAE \citep{esmaeili2019structured} (72.7) and IntroVAE \citep{huang2018introvae} (73.1) will yield worse results than vanilla VAE (73.5), which may be attributed to the increased training difficulty.

\textbf{Choices of contrastive learning methods}.
SimCLR is denoted as “SelfContrast” in Table 4. Our proposed DCCL (73.5) turns out to outperform other representative SSL approaches: SimCLR \citep{chen2020simple} (68.9 in Tab. 4), MoCo \citep{he2020momentum} (69.7), BYOL \citep{grill2020bootstrap} (70.7), SwAV \citep{caron2020unsupervised} (71.5).

\textbf{Further justification of cross-domain contrast (CDC)}.
To further justify cross-domain contrast (CDC), we also implement a baseline using within-domain positive samples only, and the accuracy drops remarkably compared to CDC (71.8 → 70.4). In addition, we include an oracle experiment with solely cross-domain positive pairs and observe comparable performance (71.8 → 71.9). It may require careful design to make good use of domain information to obtain improvements.

\textbf{Choices of pre-trained backbone and resources.}
In Table~\ref{tab:source}, we present additional experiments on Instagram (3.6B) pre-trained RegNet \cite{singh2022revisiting} and CLIP (400M) pre-trained ViT \cite{dosovitskiy2020image}. Compared to PCL, which ignores the pre-trained information, DCCL achieves consistent and substantial improvement on imagenet pre-trained models. And when applied to Instagram and CLIP, the improvement becomes remarkably larger. These indicate the importance of the pre-trained information, and more abundant the pre-training resources, the stronger the pre-trained information is needed.

\begin{table}[h]
\centering
\caption{Performance with different pre-trained resources.}\label{tab:source}
\scalebox{0.8}{
\begin{tabular}{ccccc}
\toprule
Backbone & ResNet-18        & ResNet-50        & RegNet  & \tx{ViT}         \\ 
         Resource         & \multicolumn{2}{c}{ImageNet (1.3M)} & Instagram (3.6B) & \tx{CLIP (400M)} \\ \midrule
PCL               & 65.0             & 71.6             & 73.2 & \tx{75.5}             \\ 
DCCL              & \textbf{67.5} (+2.5)      & \textbf{73.5} (+1.9)      & \textbf{82.5} (+9.3) & \tx{\textbf{78.9} (+3.4)}  \\ \bottomrule   
\end{tabular}}
\end{table}

\textbf{Further Experiments on the Wilds Benchmark.}

We also test the OOD performance of our proposed DCCL using the Camelyon and iWildCam datasets from the Wilds benchmark with the pre-trained ResNet-50 network. In Table \ref{tab:wilds}, DCCL demonstrate a consistent and substantial improvement in performance on the more challenging datasets.

\begin{table}[h]
\centering
\caption{Performance on Wilds datasets with pre-trained ResNet-50.}\label{tab:wilds}
\scalebox{1.0}{
\begin{tabular}{cccc}
\toprule
Datasets     & \multicolumn{2}{c}{Camelyon} & iWildCam \\
Metrics     & Avg. Acc     & Worst Acc     & F1       \\ \midrule
ERM  & 88.7        & 68.3         & 31.3    \\
PCL  & 91.2        & 75.5         & 30.2    \\
DCCL & \textbf{96.7}        & \textbf{90.9 }        &\textbf{32.7}   \\ \bottomrule
\end{tabular}}
\end{table}

\textbf{Further Ablation Study on the VLCS dataset.}

\textcolor{black}{Here we additionally performed an ablation study on the VLCS dataset, as shown in Table \ref{tab:vlcs_ablate}, where the performance gain above SWAD is relatively smaller. These results further confirm that the three components we identified contribute consistently to the effectiveness, as detailed in our paper.}

\begin{table}[h]
\centering
\caption{Further ablation Study on VLCS dataset with pre-trained ResNet-50.}\label{tab:vlcs_ablate}
\begin{tabular}{cccccc}
\toprule
Algorithm    & C             & L             & S             & V             & Avg           \\ \midrule
SWAD         & 98.8          & 63.3          & 75.3          & 79.2          & 79.1          \\
DCCL w/o CDC & 98.9          & 63.8          & 75.6          & 79.5          & 79.4          \\
DCCL w/o PMA & 98.6          & 63.7          & 75.7          & 79.3          & 79.3          \\
DCCL w/o GT  & 98.7          & \textbf{64.3} & 75.2          & 80.2          & 79.6          \\ \midrule
DCCL         & \textbf{99.1} & 64.0          & \textbf{76.1} & \textbf{80.7} & \textbf{80.0} \\ \bottomrule
\end{tabular}
\end{table}

\tx{\textbf{Additional Validation on Aggresive Augmentation.}}

\tx{In Table \ref{tab:ablation} of the paper, we've presented an ablation study on aggressive augmentation. Previous works usually adopted random cropping, grayscale, horizontal flipping and random color jittering. In this paper, we simply increase the intensity of random color jittering to achieve more aggressive data augmentation on all datasets. Here, we provide additional validation in Table \ref{tab:aug_ablate} by showcasing the performance of ERM and our DDCL on the OfficeHome dataset under various augmentation scenarios: without augmentation, with standard augmentation, and with aggressive augmentation. Notably, aggressive augmentation proves advantageous for our DDCL while detrimental to ERM compared to standard augmentation. Stronger and more adaptive augmentation methods for contrastive learning on DG will be explored to further enhance the performance in the future.}

\begin{table}[h]
\centering
\caption{Comparison of ERM and DCCL with different augmentation strategies.}
\label{tab:aug_ablate}
\begin{tabular}{cccccc}
\hline
 & A & C & P & R & Avg \\
\hline
ERM w/o aug & 60.2 & 52.1 & 75.6 & 78.0 & 66.5 \\
ERM w standard & 63.1 & 51.9 & 77.2 & 78.1 & 67.6 \\
ERM w aggressive & 61.7 & 51.6 & 76.3 & 77.5 & 66.8 \\
DCCL w/o aug & 66.6 & 56.9 & 81.3 & 82.1 & 71.7 \\
DCCL w standard & 69.8 & 58.6 & 81.0 & 82.6 & 73.0 \\
DCCL w aggressive & 70.1 & 59.1 & 81.4 & 83.4 & 73.5 \\
\hline
\end{tabular}
\end{table}





\subsection{\tx{Efficiency and Computation Cost}}
\label{app:efficiency}

\tx{The algorithmic complexity of our method and its baselines is complex due to factors like Feature Extraction time and Loss Calculation time. Feature extraction is consistent across all baselines, including ERM, and is a significant part. For the loss calculation, given a batch size of and a hidden dimension, and using contrastive loss calculated over batch pairs, the complexity is $O(B^2D)$, which is uniform across all contrastive learning methods.}

\tx{In this section, we present comparisons of running time (average training time per optimization step with batch\_size$=32$ and n\_steps$=5000$) and memory consumption in Table \ref{tab:comparison} among the methods. We note that our paper exhibits similar or even less time and memory costs compared to ERM and other baseline methods.}

\begin{table}[h]
\centering
\caption{Time and Memory Comparison.}
\begin{tabular}{ccc}
\hline
 & Time (s) & Memory (MiB) \\
\hline
ERM & 0.664 & 11399 \\
PCL & 0.812 & 14655 \\
SAGM & 1.326 & 12321 \\
DCCL & 0.711 & 12993 \\
\hline
\end{tabular}
\label{tab:comparison}
\end{table}

\subsection{\tx{Hyper-parameter Study}}
\label{app:hyper}

\tx{We present ablation studies on the trade-off hyper-parameters $\lambda$ and $\beta$ in Table \ref{tab:lambda_results} and \ref{tab:beta_results}. The results indicate our proposed method is stable in a wide range of hyper-parameter values. Across all selections of hyperparameters, our method stably outperforms the strongest baselines MIRO (with Avg. Acc 72.4\%).}

\begin{table}[h]
    \centering
    \caption{Results for different values of $\lambda$.}
    \begin{tabular}{cccccc}
        \hline
        $\lambda$ & A & C & P & R & Avg \\
        \hline
        0.1 & 69.7 & 59.0 & 81.4 & 83.1 & 73.3 \\
        1 & 70.1 & 59.1 & 81.4 & 83.4 & 73.5 \\
        5 & 70.3 & 58.2 & 80.9 & 83.0 & 73.1 \\
        \hline
    \end{tabular}
    
    \label{tab:lambda_results}
\end{table}

\begin{table}[h]
    \centering
    \caption{Results for different values of $\beta$.}
    \begin{tabular}{cccccc}
        \hline
        $\beta$ & A & C & P & R & Avg \\
        \hline
        0.01 & 69.8 & 59.1 & 81.0 & 82.5 & 73.1 \\
        0.05 & 70.1 & 59.1 & 81.4 & 83.4 & 73.5 \\
        0.1 & 69.5 & 58.4 & 81.4 & 83.5 & 73.2 \\
        \hline
    \end{tabular}
    
    \label{tab:beta_results}
\end{table}

\section{Discussions \& Limitations}
\label{sec:discussions}
In the paper, We analyze the failure of directly applying SCL to DG with the CL theory and suggest lack of \textcolor{black}{intra-class connectivity} in the DG setting causes the deficiency. 
We accordingly propose domain-connecting contrastive learning (\model) to enhance the connectivity across domains and obtain generalizable and transferable representation for DG. Extensive experiments also verify the effectiveness of our method. 

However, we're also aware of the \textbf{limitations} of our work. We don't make explicit use of the domain information. It implies if one can well leverage the domain information, better generalization performance might be obtained. 
{Moreover, similar to \cite{cha2022domain}, our proposed \model~requires the pre-trained embeddings of the samples. This existing drawback can be mitigated by generating the pre-trained embeddings in advance and storing them locally.} 
In addition, how to develop stronger and more adaptive augmentation methods for contrastive learning on DG is not explored in this paper and remains an open problem.

Regarding \textbf{attribution of existing assets}, we only utilize existing open-sourced datasets, which all can be found in DomainBed\footnote{\url{https://github.com/facebookresearch/DomainBed}} benchmark. In addition, we don't make any use of \textbf{personal data}. For all the datasets used, there is no private personally identifiable information or offensive content.

\section{Analysis on Intra-class Connectivity}
\label{app:analysis}

In this section, we analyze how intra-class connectivity contributes to reducing the intra-class variance based on  the concept of sample connectivity proposed in \citet{wang2022chaos}.


\begin{definition}[Sample Connectivity \cite{wang2022chaos}] Given a collection of augmentations \( A = \{ a \ | \ a : \mathbb{R}^d \rightarrow \mathbb{R}^d \} \), we say that two different samples \( x_i, x_j \in \mathbb{R}^d \) are \( A \)-connected if they have overlapped views: $\text{supp}(p(x_i^+ | x_i)) \cap \text{supp}(p(x_j^+ | x_j)) \neq \emptyset$, \text{ or equivalently, } $\exists a_i, a_j \in A \text{ such that } a_i(x_i) = a_j(x_j)$.
    
\end{definition}

Then an \textbf{augmentation graph} can be defined based on the sample connectivity. The $N$ natural samples are denoted as the vertices of the graph, and there exists an edge between two samples if they are \( A \)-connected.
The intuitive graph-based measure to assess the intra-class connectivity we previously described in \Cref{app:connectivity} is indeed motivated by the concept above of ``augmentation graph''.
In the theoretical analysis, \citet{wang2022chaos} turned to leverage a stronger condition:

\begin{assumption}[Strong Intra-class Connectivity] Given a training set \( D_s \), there exists an appropriate augmentation set \( A \) such that the augmentation graph is class-wise connected, i.e., \( \forall y \in Y \), the graph \( G_y \) restricted to vertices in class \( y \)) is connected.
    
\end{assumption}

Furthermore, they assume the perfect alignment for the minimizer of the InfoNCE (contrastive) loss:


\begin{assumption}[Perfect Alignment]
At the minimizer \( f^* \) of the InfoNCE (contrastive) loss, we can achieve perfect alignment, i.e., \( \forall x, x^+ \sim p(x, x^+) \), \( f(x) = f^*(x^+). \)    
\end{assumption}

They then attain the desired zero intra-class variance in the following proposition.

\noindent \begin{proposition}
\label{prop:invariance}
Under Assumptions 1 \& 2, by minimizing the InfoNCE loss we can conclude that the conditional variance terms vanish at the minimizer \( f^* \), i.e.,
\[
\mathrm{Var}(f^*(x) | y) = 0.
\]
\end{proposition}


Although it is impracticable to have both Assumptions 4.5 \& 4.6 hold for real-world domain generalization, we conclude from the analysis that if we can manage to increase the intra-class connectivity in SCL, the intra-class variance will accordingly shrink and benefit the consequent generalization performance.

\setcounter{page}{12}

\end{document}